\documentclass[runningheads]{llncs}

\usepackage{eccv}

\usepackage{eccvabbrv}
\usepackage{graphicx}
\usepackage{booktabs}

\usepackage[accsupp]{axessibility}  

\usepackage{hyperref}

\usepackage{orcidlink}

\begin{document}

\title{Toward Tiny and High-quality Facial Makeup with Data Amplify Learning} 

\titlerunning{Toward Tiny and High-quality Facial Makeup with Data Amplify Learning}

\author{Qiaoqiao Jin\inst{1}\thanks{This work was completed during the author's internship at Alibaba.} \and
Xuanhong Chen\inst{1,3} \and
Meiguang Jin\inst{2}\and
Ying Chen\inst{2} \and
Rui Shi\inst{1} \and
Yucheng Zheng\inst{1} \and
Yupeng Zhu\inst{1} \and
Bingbing Ni\inst{1}\thanks{Corresponding author: Bingbing Ni.}
}

\authorrunning{Q.Jin et al.}

\institute{
{\textsuperscript{1}Shanghai Jiao Tong University, Shanghai 200240, China}\\
{\textsuperscript{2}Alibaba Group} \,{\textsuperscript{3}Institute of Cultural and Creative Industry, USC-SJTU}\\
\email{\{jinqiaoqiao, nibingbing\}@sjtu.edu.cn}\\
}




\maketitle

\begin{abstract}
Contemporary makeup approaches primarily hinge on unpaired learning paradigms, yet they grapple with the challenges of inaccurate supervision (e.g., face misalignment) and sophisticated facial prompts (including face parsing, and landmark detection).
These challenges prohibit low-cost deployment of facial makeup models, especially on mobile devices.
To solve above problems, we propose a brand-new learning paradigm, termed "Data Amplify Learning (DAL)," alongside a compact makeup model named "TinyBeauty."
The core idea of DAL lies in employing a Diffusion-based Data Amplifier (DDA) to "amplify" limited images for the model training, thereby enabling accurate pixel-to-pixel supervision with merely a handful of annotations.
Two pivotal innovations in DDA facilitate the above training approach:
(1) A Residual Diffusion Model (RDM) is designed to generate high-fidelity detail and circumvent the detail vanishing problem in the vanilla diffusion models;
(2) A Fine-Grained Makeup Module (FGMM) is proposed to achieve precise makeup control and combination while retaining face identity.
Coupled with DAL, TinyBeauty necessitates merely $\textbf{80K}$ parameters to achieve a state-of-the-art performance without intricate face prompts.
Meanwhile, TinyBeauty achieves a remarkable inference speed of up to \textbf{460 fps} on the iPhone 13.
Extensive experiments show that DAL can produce highly competitive makeup models using only \textbf{5} image pairs. Please visit \url{https://github.com/TinyBeauty} for code and demos.

  \keywords{Facial Makeup \and Image 
Synthesis \and Stable Diffusion}
\end{abstract}

\section{Introduction}
\label{sec:intro}

Facial makeup aims to enhance facial appearance by applying cosmetic components on facial images, such as vibrant lipstick, intense eyeshadow, and eyeliner.
Its primary application scenarios revolve around mobile devices, including short videos and live broadcasts, which makes makeup a time- and resource-sensitive task that demands efficient solutions. 
Specifically, mobile deployment faces unique challenges due to stringent restrictions on computational resources, which require extremely small model sizes (e.g., $<$100K) and minimized inference latency (typically $<$30ms) to work within such constrained resources.
However, the current advanced makeup methods~\cite{BeautyREC, BeautyGAN, Pix2Pix, PSGAN, SCGAN, SOGAN, ELEGANT} suffer from large model sizes, generally exceeding 1M, and heavily rely on complex face prompts pipelines to ensure accurate makeup transfer. 
These factors hinder the practical application of existing approaches.

\begin{figure}[t]
  \centering
   \includegraphics[width=1.0\linewidth]{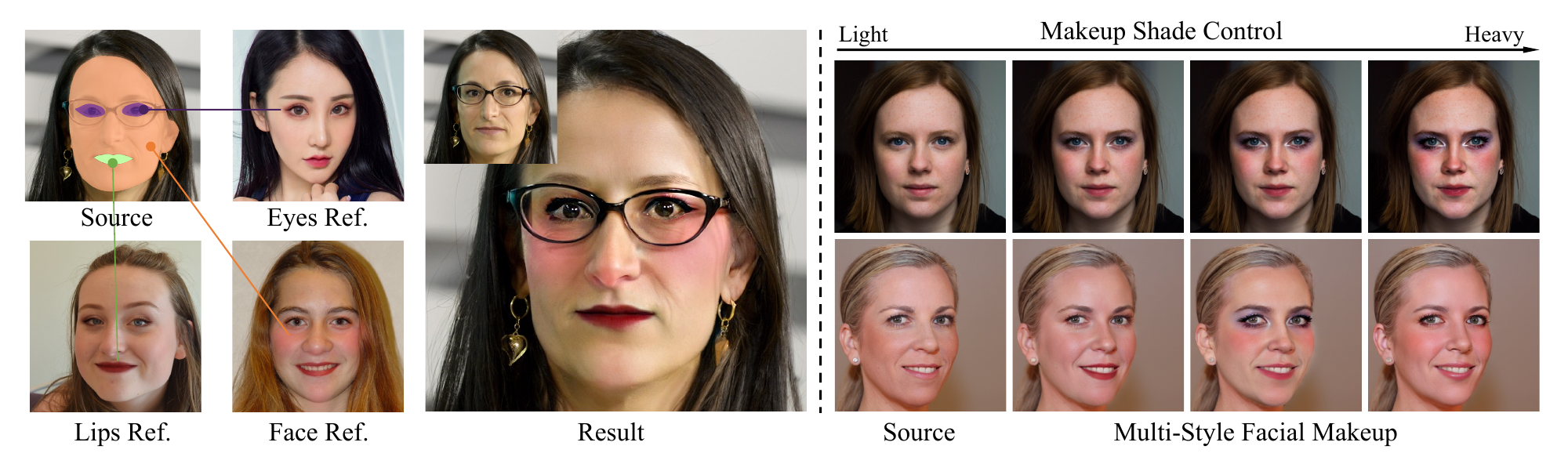}
   \caption{Our \textbf{Diffusion-based Data Amplifier} is capable of learning from just several images to generate high-quality makeup visuals in diverse styles, offering flexible control such as makeup shade control and customized local editing.}
   \label{fig:top}
   \vspace{-0.3cm}
\end{figure}
\begin{figure}[t]
  \centering
   \includegraphics[width=1.0\linewidth]{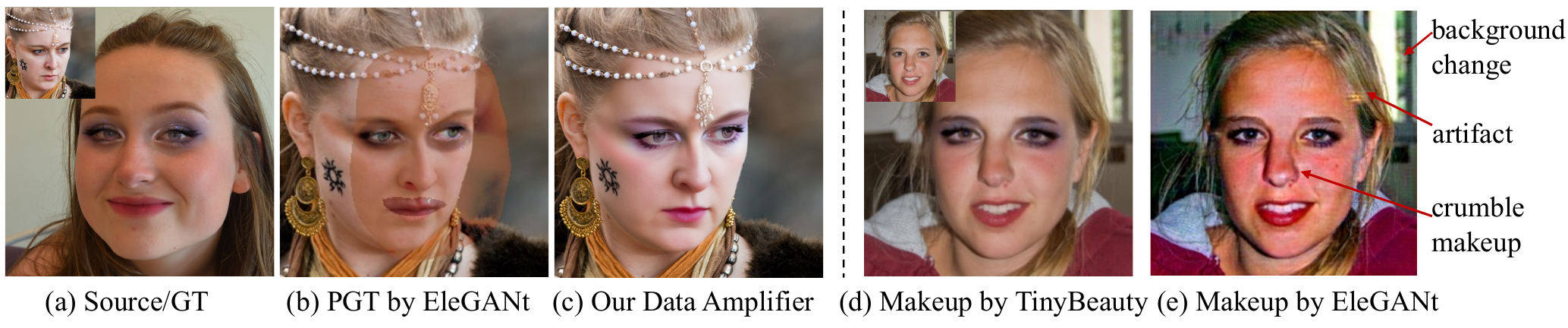}
   \caption{\textbf{Left}: Data evolution in facial makeup: unpaired data$\rightarrow$pseudo-paired data$\rightarrow$paired data (generated by data amplifier). \textbf{Right}: Comparison between TinyBeauty trained on paired data and EleGANt~\cite{ELEGANT} trained on pseudo-paired data.}
   \vspace{-0.3cm}
   \label{fig:data}
\end{figure}

The root cause behind the excessive model sizes and complex inference pipelines in current mainstream makeup models can be attributed to the underlying learning framework, which is inherently flawed.
The annotation of paired makeup data is expensive and challenging to achieve style scalability. 
Consequently, unpaired data has become the mainstream data protocol, which requires makeup models to carefully employ adversarial training~\cite{BeautyGAN,SCGAN,PSGAN} for achieving stable unpaired learning.
Moreover, unpaired data often suffers from significant facial misalignment issues, which necessitates the inclusion of facial spatial prompts and other operations (e.g., warping) to assist in the unpaired learning process. 
For example, explicit incorporation of facial landmarks is used to guide the network in identifying the position of lipstick, eyelashes, and eyeshadow. 
However, these additional prompts further contribute to the complexity of the model.
Simultaneously, the data misalignment poses a challenge for unpaired learning models to employ stable and straightforward reconstruction loss (e.g., L1, L2). Instead, they rely on inaccurate supervision, such as color histograms matching~\cite{BeautyGAN}, earth mover's distance, due to the lack of alignment information.
As well known, these inaccurate supervision methods often rely on theoretical approximations (e.g., approximate upper and lower bounds), leading to poor model robustness.
For instance, as shown in Fig.~\ref{fig:data}, even with a model size of 10M, EleGANt~\cite{ELEGANT} still frequently produces unsatisfactory results.
The above challenges become more pronounced as the model size continues to decrease.
The reduced model scale makes the model more sensitive to the accuracy of supervision while also requiring the abandonment of various auxiliary alignment techniques.

To address the above challenges, we propose a novel learning framework named \textbf{Data Amplify Learning} enabling stable and robust training supervision even with a limited number of image annotations.
In the heart of DAL is a meticulously designed \textbf{Diffusion-based Data Amplifier}, responsible for amplifying the limited labeled image pairs (typically 5 pairs) into a larger dataset suitable for model training.
This process allows DAL to transform the inaccurate supervision into pixel-to-pixel learning (e.g., L1), reducing the optimization difficulty and enabling accurate gradient propagation.
Naively using mainstream diffusion-based control methods~\cite{IP-adapter,StableDiffusion} as the Data Amplifier can lead to subpar outcomes, including identity mismatching, over-smoothing details, and inaccurate makeup results.
To alleviate above challenges, two ingredients of the DDA are proposed:
(1) Instead of directly generating makeup results, we design a Residual Diffusion Model (RDM) for the Data Amplifier. 
RDM consists of two branches: one is responsible for face reconstruction, and the other is dedicated to applying makeup. 
By utilizing the residual between the reconstructed face and the input face, we can transfer this residual to the makeup branch, resulting in high-fidelity and sharpened makeup results.
(2) Unlike direct image injection~\cite{IP-adapter}, we propose a Fine-Grained Makeup Module (FGMM) that couples fine-grained makeup with facial identity representation to control makeup process, which effectively avoids the loss of facial identity. 
Additionally, we introduce semantic region labels for makeup, further refining the control of makeup application, rendering a precise makeup control generation.
DAL greatly relaxes the optimization methods, allowing us to abandon face prompts and over-parameterization methods, thereby focusing on network design for mobile computing.
Benefiting from this, a single \textbf{TinyBeauty} model is designed without reliance on external face prompts. 
TinyBeauty exceeds prior makeup transfer performance using only 14 convolution layers, enabling efficient deployment on mobile devices. 
Furthermore, to produce more complete makeup components compared to prior works, we utilize an edge operator to constrain the learning of eyeliner, enhancing the realism and fidelity of generated makeup styles.

Quantitatively, TinyBeauty outperforms EleGANt~\cite{ELEGANT} with a significant $+5.21$dB PSNR increase (17.3\% improvement) on the FFHQ~\cite{FFHQ} dataset and $+1.49$dB (4.55\% improvement) on the MT~\cite{BeautyGAN} dataset. It also operates $13\times$ faster than BeautyGAN~\cite{BeautyGAN} and delivers a swift 2.18ms latency, ensuring smooth smartphone integration on an iPhone13.
Extensive experimental results demonstrate that DAL achieves impressive performance in training makeup models, even with a minimal dataset of only 5 image pairs.


\section{Related Work}
\label{sec:related}
\subsection{Diffusion-based Image-to-Image Translation}
Compared to previously widely used GAN-based~\cite{GAN, DBLP:conf/mm/ChenCNG20, DBLP:conf/eccv/ChenNLLJTT20, DBLP:conf/mm/ChenYLQN20, DBLP:journals/pami/ChenNLLZW24,DBLP:conf/mmm/QiuNLC21} models 
, recent diffusion models~\cite{song2020denoising,dhariwal2021diffusion,saharia2022palette} have demonstrated superior performance in high-fidelity image generation tasks, especially when few-shot images are available. Typically, conditions are introduced to steer the image generation process, generally in two forms: text-based and image-based guidance. DALL·E2~\cite{nichol2021glide} and Imagen~\cite{saharia2022photorealistic} apply text encoders from pre-trained language models~\cite{radford2021learning,devlin2018bert} for image generation guidance. Textual Inversion\cite{gal2022image} and DreamBooth\cite{ruiz2023dreambooth} learn special tokens from example images. However, solely using text for conditional guidance is indirect and inaccurate, making it difficult to ensure stable, consistently generated images. To strengthen conditional constraints, SD Image Variations2~\cite{SD1} and Stable unCLIP3~\cite{SD2} directly fine-tune text-conditioned diffusion models on image embeddings. IP-Adapter~\cite{IP-adapter} and InstantID~\cite{wang2024instantid} use a combined image and text prompt for data generation conditions to keep the identity of the condition image. While these methods yield commendable outcomes in image stylization, they fail to preserve the facial detail like wrinkles of the original image, thus inadequate in applying facial makeup. 

\begin{figure*}[t]
  \centering
   \includegraphics[width=1\linewidth]{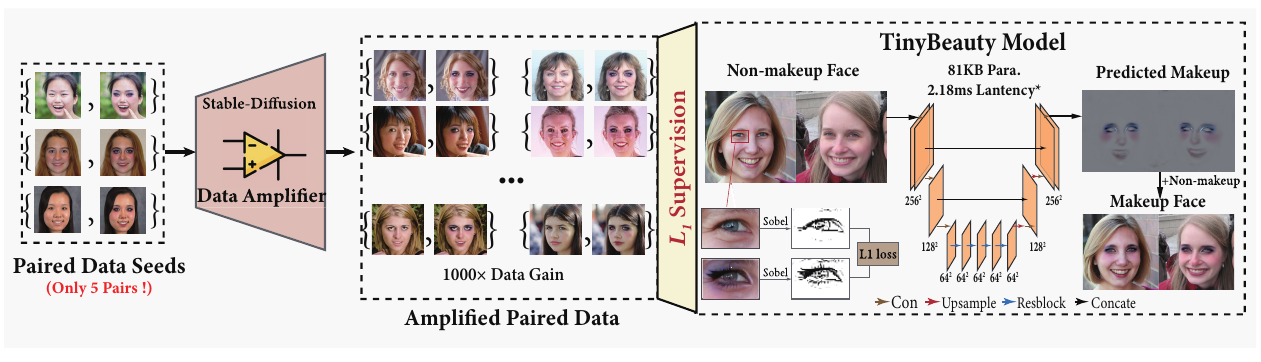}
   \caption{\textbf{Overview of the Data Amplify Learning framework.} The Data Amplify Learning process contains two components: \textbf{(1)A data amplifier} which utilizes a pretrained diffusion model to amplify a small set of seed data into a larger synthesized dataset. \textbf{(2) A lightweight model} which is trained on the amplified data to accurately learn the makeup styles while retaining identity features of the original images. \\
   $^*$The latency is the inference time on an iPhone 13 device.}
   \label{fig:pipe}
   \vspace{-0.3cm}
   
\end{figure*}

\subsection{Facial Makeup}
Traditional facial makeup methods~\cite{DBLP:conf/cvpr/GuoS09, DBLP:conf/cvpr/LiZL15} rely on facial landmarks to warp predetermined beauty materials on the facial images for makeup application, which is efficient but unrealistic.
BeautyGAN~\cite{BeautyGAN} introduces a dual GAN approach for makeup transfer with a color histogram loss using the proposed dataset. PSGAN~\cite{PSGAN} and FAT~\cite{FAT} propose attention mechanisms to address pose/expression changes, significantly increasing model size. SCGAN~\cite{SCGAN} encodes styles into component-wise codes while EleGANt~\cite{ELEGANT} simplifies this complex optimization problem as L1 loss by generating pseudo ground truth. However, EleGANt still consumes significant computational overhead due to its use of multi-scale attention modules. By abandoning CycleGAN's structure, BeautyREC~\cite{BeautyREC} achieves a lighter model compared to previous work. Unlike makeup transfer methods, the face beautification proposed in~\cite{FaceBeautification} targets many-to-many translation by integrating beauty score prediction.Typically, all these methods incorporate parsing maps and/or landmarks during pre-processing on the unpaired MT dataset, forming cumbersome pipelines.

\section{Method}
\label{sec:method}

We propose a novel learning scheme called \textbf{Data Amplify Learning (DAL)} to replace the unstable and inaccurate unpaired learning. 
DAL leverages a diffusion model to generate previously inaccessible paired data, which is then employed as training material for a compact tiny model. 
The core of DAL is a \textbf{Diffusion-based Data Amplifier (DDA)} to generate large amounts of paired data using only 5 makeup images as data seeds.~(\cref{sec:3.3}). Benefit from the DDA-amplified data, a single \textbf{TinyBeauty Model} is designed to replace the previous cumbersome makeup pipeline, facilitating the integration of the makeup model into mobile devices.~(\cref{sec:3.4}).

\begin{figure*}[t]
  \centering
   \includegraphics[width=0.98\linewidth]{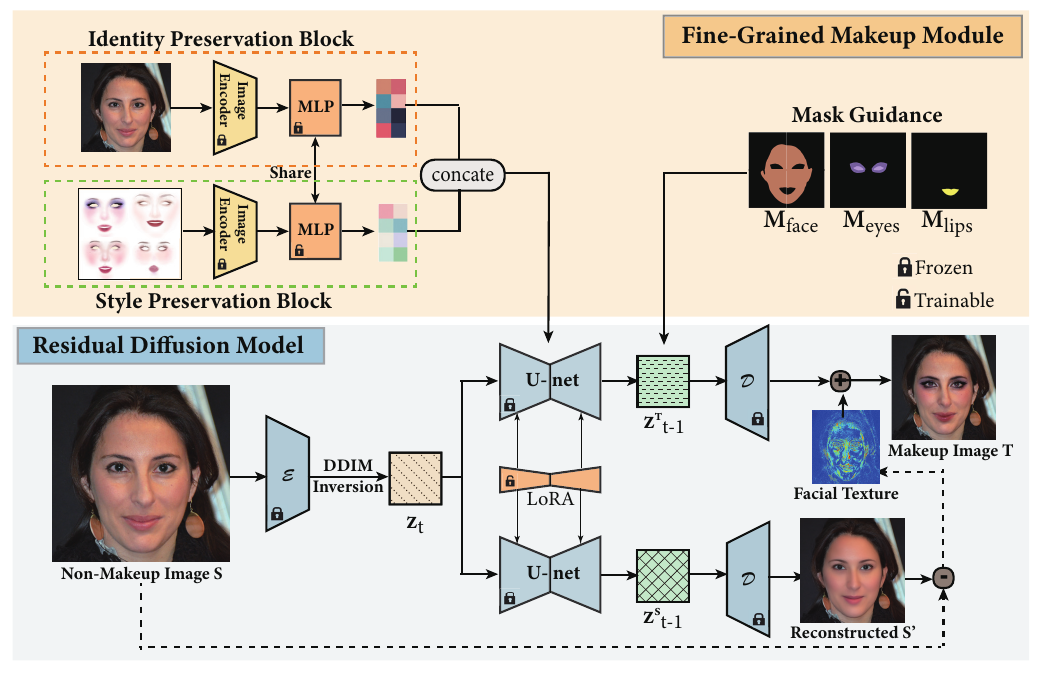}
   \caption{\textbf{Overview of the Diffusion-based Data Amplifier (DDA).} Our DDA leverages a Residual Diffusion Model for high-fidelity texture preservation, minimizing distortion and avoiding unnatural mask-like appearances. It also employs a Fine-Grained Makeup Module including Identity Preservation Block (IPB) to maintain the original facial features, Style Preservation Block (SPB) to guarantee consistent makeup style application, and facial masks to specify makeup region.}
   
   \label{fig:diffusion_main}
\end{figure*}

\subsection{Diffusion-based Data Amplifier}
\label{sec:3.3}
Diffusion models~\cite{SD, DDPM} are probabilistic generative models trained to learn a data distribution by performing an iterative denoising process. To project non-makeup image domain $X$ into makeup domain $Y$, we finetune a pre-trained Stable Diffusion (SD) model~\cite{StableDiffusion} $\mathcal{F}_{init}$ using low-rank adaption~\cite{LoRA}. We aim to guide the model to apply makeup style on input non-makeup data under conditional constraints. The constraint of the finetuning process of the SD model $\epsilon_{\theta}$ is defined as:
\begin{equation}
    L_{simple} = \mathbb{E}_{\mathbf{z}_0, \epsilon\sim\mathcal{N}(\mathbf{0},\mathbf{I}), \mathbf{c}, t}||\epsilon - \epsilon_{\theta}(\mathbf{z}_t, \mathbf{c}, t)||^2, 
\end{equation}
where $\mathbf{z}_0$ represents the latent feature of the manually annotated makeup image with condition $\mathbf{c}$, $\epsilon$ denotes the noise added to $\mathbf{z}_0$, $t \in [0, T]$ denotes the time step of diffusion process, $\mathbf{z}_t = \alpha_t \mathbf{z}_0 + \delta_t \epsilon$ is the noisy data at $t$ step. To generate makeup images, we incorporate conditions to guide the inference process in the fine-tuned SD model $\mathcal{F}_{fine}$.
\begin{equation}
    y =\mathcal{F}_{fine}(x, \mathbf{c}),
\end{equation}
where x is the non-makeup image, and y is the paired makeup image.

To generate high-quality paired makeup data, our DDA is required to contain the subject's original facial features, skin texture details, such as wrinkles and spots, and consistent, precise makeup styles across various portraits. However, as shown by Gal et al.~\cite{gal2022image}, stable diffusion models face a trade-off between image reconstruction and editability, making it difficult to apply makeup while preserving the original face. To overcome these obstacles, we introduce a Residual Diffusion Model that preserves texture and detail, reducing distortion and mask-like effects. Moreover, we propose a Fine-Grained Makeup Module to ensure the precise application of makeup to the appropriate facial areas and generate visually consistent makeup styles, as shown in~\cref{fig:diffusion_main}. 

\noindent\textbf{Residual Diffusion Models}
To address the issue of facial features like wrinkles being smoothed out during the denoising step of diffusion-based portrait image generation, we introduce a novel framework that utilizes parallel diffusion branches during the inference phase—the conditioned and unconditioned branches. The conditioned branch operates with content condition $\mathbf{c}_{con}$ and style condition $\mathbf{c}_{sty}$ to produce a smooth makeup image, while unconditioned branch proceeds without any conditions to output a smooth non-makeup image.

We define the concept of residual detail $R_{d}$ by considering the difference between the original image $x$ and its smooth counterpart produced by the unconditioned branch, $\mathcal{F}_{fine}(x)$. This difference, $R_{d} = x - \mathcal{F}_{fine}(x)$, represents the facial detail residuals, which are the essential features we aim to preserve to maintain realism.
Furthermore, the makeup residual $R_{m}$ is captured by the difference between the outputs of the two branches, $R_m = \mathcal{F}_{fine}(x, (\textbf{c}_{sty} + \textbf{c}_{con})) - \mathcal{F}_{fine}(x)$. 
To synthesize the enhanced portrait, we apply the following equation:
\begin{equation}
y_{detail} = \mathcal{F}_{fine}(x) + \lambda_m R_m + \lambda_d R_d,
\end{equation}
where controlling the coefficients of the makeup $\lambda_m$ and detail residuals $\lambda_d$ allows us to adjust the makeup intensity and detail sharpness, respectively. Usually, we set $\lambda_m$ to 1 and set $\lambda_d$ to 0.8. This control mechanism is akin to an annealing concept, where increasing $\lambda_d$ to 1.0 can lead to an oversaturation effect due to the addition of makeup-related details.
This framework not only improves the quality of makeup in generated images but also has broader applications in other high-fidelity image generation tasks involving diffusion models, offering an advanced technique for creating realistic and detailed synthesized images.

\noindent\textbf{Fine-Grained Makeup Module}
The Fine-Grained Makeup Module comprises the Identity Preservation Block (IPB), which safeguards individual facial identity, the Style Preservation Block (SPB) for precise makeup style management, and a latent face mask that guides region-specific makeup application and facilitates the flawless integration of diverse makeup styles.

\textbf{Style Preservation Block (SPB)}.
The crux for DDA lies in the accurate replication of makeup styles. Traditional diffusion-based image synthesis methods~\cite{IP-adapter,wang2024instantid}, which use text to describe artistic styles, are ineffective for the nuanced task of capturing makeup. Text descriptions might broadly categorize lipstick as "red" but cannot detail the specific shade, undertone, or finish and how it interacts with the skin tone. Therefore, we employ visual examples as style references instead of textual references. To focus solely on the makeup while minimizing the influence of other facial features, we apply makeup to a frontal face image and employ a facial mask to isolate the makeup-applied regions, creating clean and unobstructed makeup images as style references. Then we encode the style image using a pre-trained image encoder and use a trainable MLP to project it as style tokens.

\textbf{Identity Preservation Block (IPB)}.
A critical aspect of our DDA is the intrinsic maintenance of facial identity, especially when learning makeup styles, which can inadvertently morph facial features.
To maintain facial identity, our DDA employs the Identity Preservation Module (IPB) to disentangle style from identity, ensuring that the addition of makeup styles does not alter the facial structure. We initially considered using a distinct facial encoder like ArcFace~\cite{ArcFace} for the IPB but faced challenges due to incompatible encoding spaces. Thus, we unify the encoding space for SPB and IPB, leveraging a shared MLP for feature blending. 
Then the global condition vector $\mathbf{c}$ is divided into independent content $\mathbf{c}_{con}$ and style $\mathbf{c}_{sty}$ components. 

\textbf{Mask Guidance}.
To accurately define the facial areas affected by the makeup style, we divide the feature space into three distinct regions under the guidance of resized facial masks: $M_{face}$, $M_{lips}$, and $M_{eyes}$, as shown in~\cref{fig:diffusion_main}. For single makeup style learning, we define the influenced area $M_{changed}$ as
$M_{changed} = M_{face} + M_{lips} +M_{eyes}$. In training process, we only focus the learning on $M_{changed}$ regions to effectively capture makeup style. This is achieved by computing the loss $L_{simple}$ only over the features within the $M_{changed}$ area:
\begin{equation}
    L_{simple}^M = \mathbb{E}_{\mathbf{z}_0, \epsilon\sim\mathcal{N}(\mathbf{0},\mathbf{I}), \mathbf{c}, t}||(\epsilon - \epsilon_{\theta}(\mathbf{z}_t, \mathbf{c}, t))* M_{changed}||^2 .
\end{equation}
In the inference phase, the application of the latent space is similarly restricted to the regions specified by the changed mask. The blended latent $L_y'$ is a combination of the output latent of the diffusion process $L_y$ in the makeup-altered regions and the original latent $L_x$ in the unaffected regions. This blending is mathematically represented as:: $L_y' = L_y \bigodot M_{changed} + L_x \bigodot (1-M_{changed})$. 
When learning to combine multiple makeup styles, the model needs to handle various makeup conditions simultaneously. By applying different makeup conditions to each respective facial mask, we can impose constraints that allow for a seamless integration of multiple makeup elements on a single portrait.


\subsection{TinyBeauty Model}
\label{sec:3.4}
Previous makeup pipelines are trained on unpaired data and, therefore heavily rely on face prompts like landmarks and facial masks to achieve face alignment. This reliance on heavy pipelines results in bulky networks with unstable and subpar accuracy. Benefiting from DDA-generated paired data, the TinyBeauty Model sidesteps this laborious pre-processing by directly applying L1 loss aligning generated images closely with their targets. Furthermore, we observed that traditional GAN-based methods directly generate makeup images as outputs, introducing noise to regions that should remain unaffected, like the background and hair~\cref{fig:data}. To mitigate these unwanted artifacts, we design TinyBeauty to output only the makeup residuals, as depicted in ~\cref{fig:pipe}. This innovative approach significantly diminishes artifacts and ensures the preservation of the original image's features and details.

\textbf{Network Architecture.} Benefiting from DDA-generated data, TinyBeauty can be designed as a hardware-friendly network optimized for resource-constrained devices. We construct our TinyBeauty entirely from the most basic building block - fully convolutional layers. As depicted in~\cref{fig:pipe}, we implement a U-Net-based~\cite{UNet} generator architecture containing only four convolutional layers and four residual blocks, comprising a total of fourteen convolutional operations. This architectural choice is highly efficient while retaining the ability to capture spatial dependencies. 
When hiring the lightweight network, another key advantage of TinyBeauty is that the outputted residuals can be applied to images of any resolution without losing the original texture information.
As a result, our tiny makeup model maintains the capability to translate fine-grained makeup details with a compact network architecture comprising only 81KB parameters.

\textbf{Eyeliner Loss.}
Previous methods~\cite{BeautyGAN, ELEGANT, BeautyREC, SCGAN} fail to learn clear eyeliner which holds pivotal significance in makeup, due to using unpaired data that hampers the network's ability to match and learn such fine details. To enhance the delineation of eyeliners, which typically exhibit evident edges, we employ edge filters for their extraction. We specifically use the Sobel filter $\mathcal{S}$ to extract the edges effectively from the eyeliner area. These extracted edges are then leveraged by a specialized loss function $\mathcal{L}_s$ to accurately define the eyeliner's contours.
\begin{equation}
    \mathcal{L}_s = ||\mathcal{S}(y) - \mathcal{S}(y')||_2^2 * M_{eyes},
\end{equation}
where $y$ is makeup image, $y' = M(x) + x$ is the predicted makeup image, $M_{eyes}$ is the mask around eyes.

\textbf{Reconstruction Loss.} 
Previous methods, lacking paired data, resorted to complex loss functions like histogram matching for makeup applications. In contrast, thanks to our DDA-generated paired data, we can directly use a global L1 loss to enforce constraints.
\begin{equation}
    \mathcal{L}_{rec} = || y - y'||_1.
\end{equation}

\begin{figure*}[t]
  \centering
   \includegraphics[width=1\linewidth]{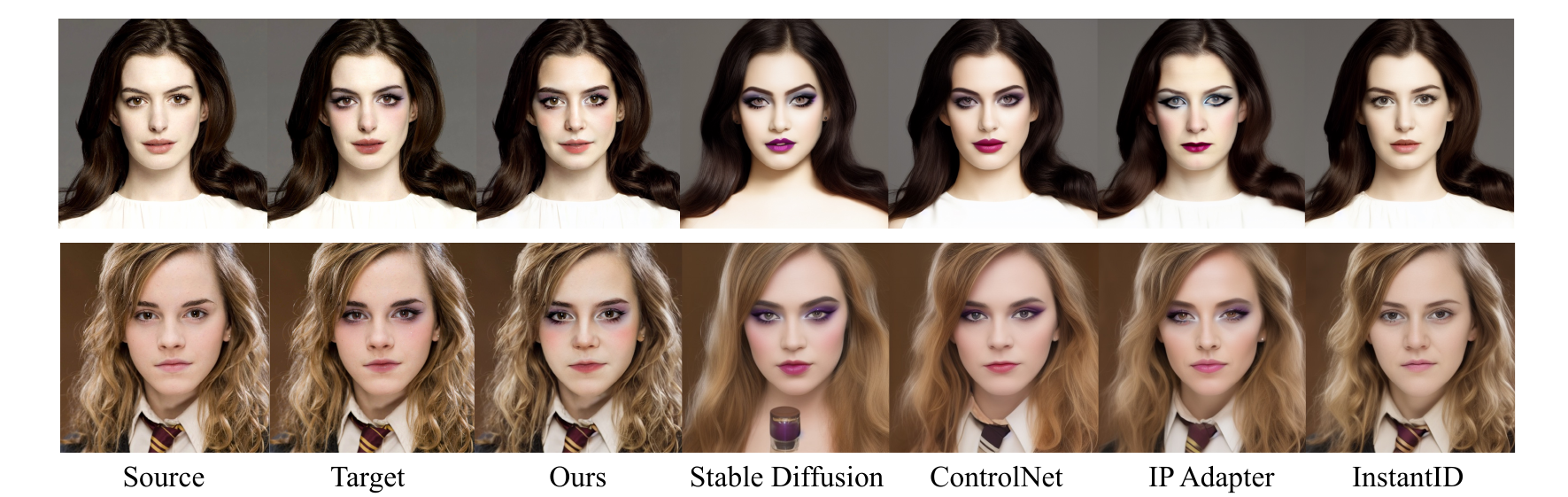}
   \caption{\textbf{Visual Comparison of Facial Makeup Using DDA.} Comparative results highlight the superior performance of DDA in maintaining facial integrity and style consistency when compared with alternative methods.}
   
   \label{fig:res-diff}
\end{figure*}
In addition to the above losses, we also leverage the perception loss $\mathcal{L}_{per}$ and adversarial loss $\mathcal{L}_{adv}$, which have been widely utilized in previous researches~\cite{BeautyGAN, ELEGANT, SCGAN, BeautyREC}. The supplementary material provides more details of these losses.

\section{Experiments}
\subsection{Datasets and Evaluation Metrics}

\textbf{Data Seeds and Evaluation Dataset.} We randomly select \emph{5} images without makeup from the Flickr-Faces-HQ (FFHQ)~\cite{FFHQ} dataset as our DAL's data seeds, \ie training data. We then manually annotate these data seeds with five predefined makeup styles using MEITU~\footnote{https://www.meitu.com/}, along with the makeup-only images as the input of SPB. 
To enable a comprehensive and fair evaluation, 100 images from the FFHQ dataset~\cite{FFHQ} and 100 images from the MT dataset~\cite{BeautyGAN} are additionally annotated as test data when computing PSNR, FID, and LPIPS.

\begin{figure*}[t]
  \centering
   \includegraphics[width=1\linewidth]{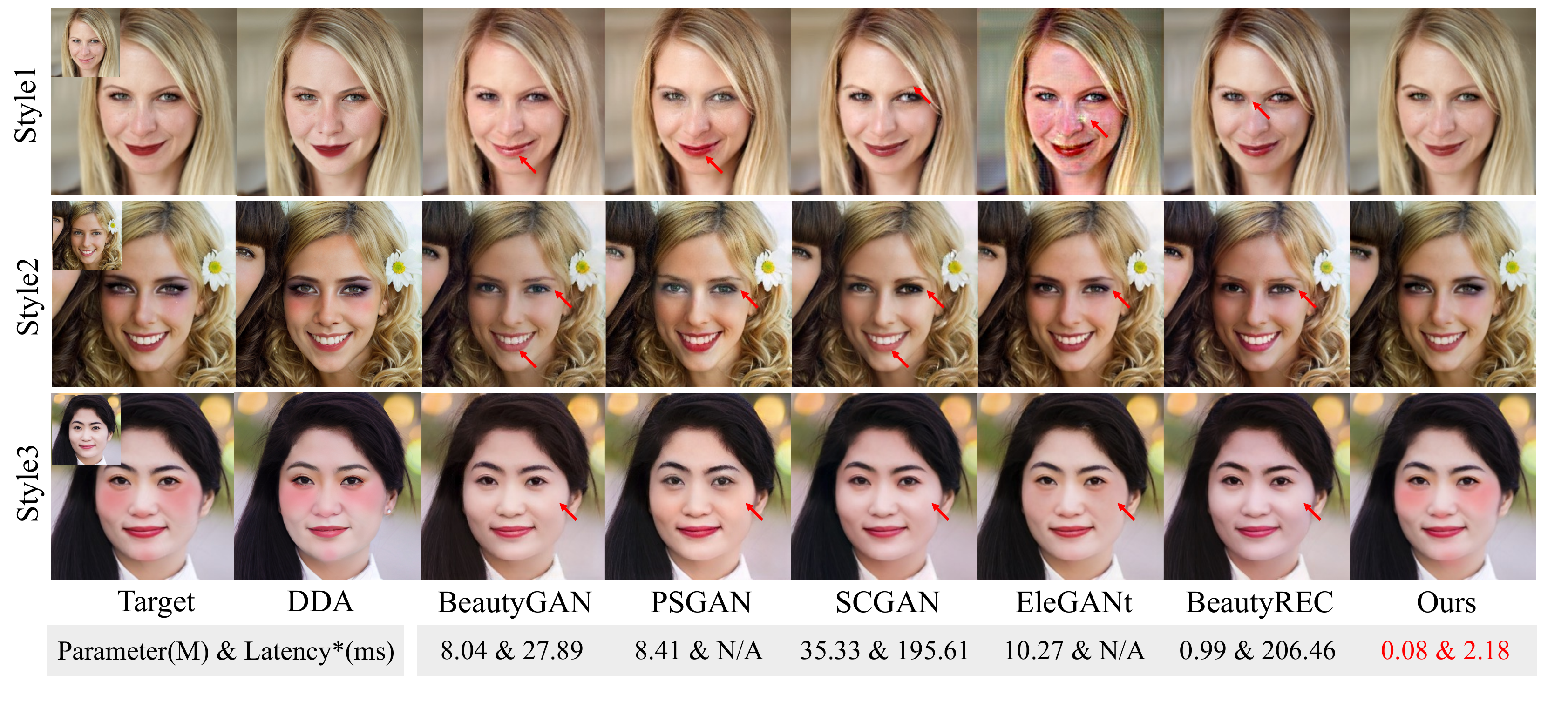}
   \caption{\textbf{Visual comparison of TineBeauty on the FFHQ~\cite{FFHQ} images,} with the parameter size and inference latency of each model \textbf{on an iPhone13 device}. N/A means the model cannot be deployed on iPhone devices. (Find more results of MT Datasets and more makeup styles in Supplementary Material.)}
   \label{fig:res}
\end{figure*}

\textbf{Evaluation Metrics.} To evaluate TinyBeauty quantitatively, we employ widely recognized metrics to evaluate the similarity, diversity, and realism of the images generated. We use the Peak Signal-to-Noise Ratio (PSNR) to measure the similarity between the generated data and ground truth (GT) data. A higher PSNR indicates a closer approximation to the GT. Additionally, we report the Fr\'{e}chet Inception Distances (FID) and Learned Perceptual Image Patch Similarity (LPIPS) by comparing 100 generated images to 100 GT images. FID and LPIPS metrics evaluate the diversity and realism of the generated samples, with lower scores indicating a higher similarity to the GT data distribution.

\subsection{Implementation Details}
\textbf{Diffusion-based Data Amplifier (DDA).} Our diffusion fine-tuning experiment use SDv1.5~\cite{diffuers} as the base model. The image encoder utilized is OpenCLIP ViT-H/14~\cite{open_clip}, which is pre-trained by IP-Adapter~\cite{IP-adapter}. We set the length of both style token and identity token to 32 to balance the strength of style and identity. LoRA~\cite{LoRA} in conjunction with U-net is employed to fine-tune the SD model with a learning rate of 1e-4 for 500 epochs. Notably, the five distinct makeup styles are concurrently trained within a unified model, demonstrating the model's capacity for multi-style learning in a single training session. The training process is executed on an NVIDIA V100 GPU, taking about 50 mins to fine-tune five makeup styles. We utilize the FaRL~\cite{FaRL} to generate facial masks and resize them to a resolution of $64\times64$ to guide the network's training in the latent space.



\textbf{TinyBeauty Model.} The TinyBeauty model is trained using a dataset consisting of 4000 diffusion-generated images for 50 epochs. The training processes utilized a learning rate of $2e-4$ and the Adam~\cite{Adam} optimizer. The entire training procedure is executed on an NVIDIA V100 GPU, requiring approximately 12 hours. It is important to mention that \textbf{NO} specific face operations are conducted on the input image during the inference phase, which makes our makeup process fast and hardware-friendly.

\begin{table}[t]
        \centering
        \caption{Results of Diffusion-based Data Amplifier (DDA), TinyBeauty, and competing methods on FFHQ~\cite{FFHQ} dataset and MT~\cite{BeautyGAN} dataset, in style1. The best and second best results of each column are indicated with bold font and underlined respectively. * means the models are trained with our proposed DAL scheme.}
        \fontsize{8}{10}\selectfont    
        \begin{tabular}{ccccccc}
                \toprule
                Method&
                \multicolumn{3}{c}{FFHQ} &
            \multicolumn{3}{c}{MT}\cr
                \cmidrule(lr){2-4}  
                \cmidrule(lr){5-7}  
                & PSNR$\uparrow$ & FID$\downarrow$ & LPIPS$\downarrow$ & PSNR$\uparrow$ & FID$\downarrow$ & LPIPS$\downarrow$ \cr
                \cmidrule(lr){1-7}

                BeautyGAN~\cite{BeautyGAN} & 26.50 & 45.25 & 0.0564 & 27.49 & 25.05 & 0.0434 \cr
                PSGAN~\cite{PSGAN}     & 25.65 & 36.22 & 0.0594 & 28.05 & 18.72 & 0.0301 \cr
                SCGAN~\cite{SCGAN}     & 27.55 & 36.98 & 0.0485 & 27.22 & 30.85 & 0.0467 \cr
                EleGANt~\cite{ELEGANT}   & 30.18 & 25.47 & 0.0396 & 32.77 & 12.55 & \underline{0.0191} \cr
                BeautyREC~\cite{BeautyREC} & 24.93 & 26.88 & 0.0538 & 27.39 & 21.37 & 0.0430 \cr

            \cmidrule(lr){1-7}
                EleGANt*~\cite{ELEGANT}   & \underline{35.45} & 10.78 & \underline{0.0148} & 34.65 & 11.57 & \textbf{0.0164} \cr
                BeautyREC*~\cite{BeautyREC} & 31.06 & 15.59 & 0.0374 & 27.39 & 18.21 & 0.0232 \cr
            \cmidrule(lr){1-7}   
                        \textbf{DDA} & \textbf{35.96} & \underline{10.28} & 0.0195 & \textbf{34.79} & \underline{10.37} & 0.0231 \cr
                        
                        \textbf{TinyBeauty} & 35.39 & \textbf{8.03} & \textbf{0.0146} & \underline{34.26} & \textbf{9.33} & \underline{0.0181} \cr

                \bottomrule
        \end{tabular}
        \label{tab:PSNR}
\end{table}

\subsection{Comparison}

To showcase the quality of the data generated using our DDA, we compare it with the text-conditioned SD model, ControlNet~\cite{zhang2023adding}, IP-Adapter~\cite{IP-adapter}, and InstantID~\cite{wang2024instantid}. We train both previous methods in the same data as ours, with the result shown in~\cref{fig:res-diff}. The results reveal two key insights. Firstly, the subtleties of makeup styles are not easily captured by textual descriptions alone. Consequently, it is challenging to precisely manipulate attributes, particularly color, in the synthesized images. Our DDA addresses this by directly encoding makeup images, thus producing results with consistent color and makeup content. Secondly, IP-Adapter falls short of meeting our stringent criteria for identity retention in the facial makeup task. Our DDA, however, effectively resolves this issue, generating high-quality makeup images by synergistically leveraging mask guidance, IPB, and RDM. Additionally, \cref{fig:top} illustrates the DDA's capability of fusing various makeup styles. Our DDA adeptly blends eye, lip, and blush makeup to create a myriad of diverse and intricate makeup applications.

\begin{figure}[t]
    \centering
    \begin{minipage}[t]{0.54\linewidth}
        \centering
        \includegraphics[width=\linewidth]{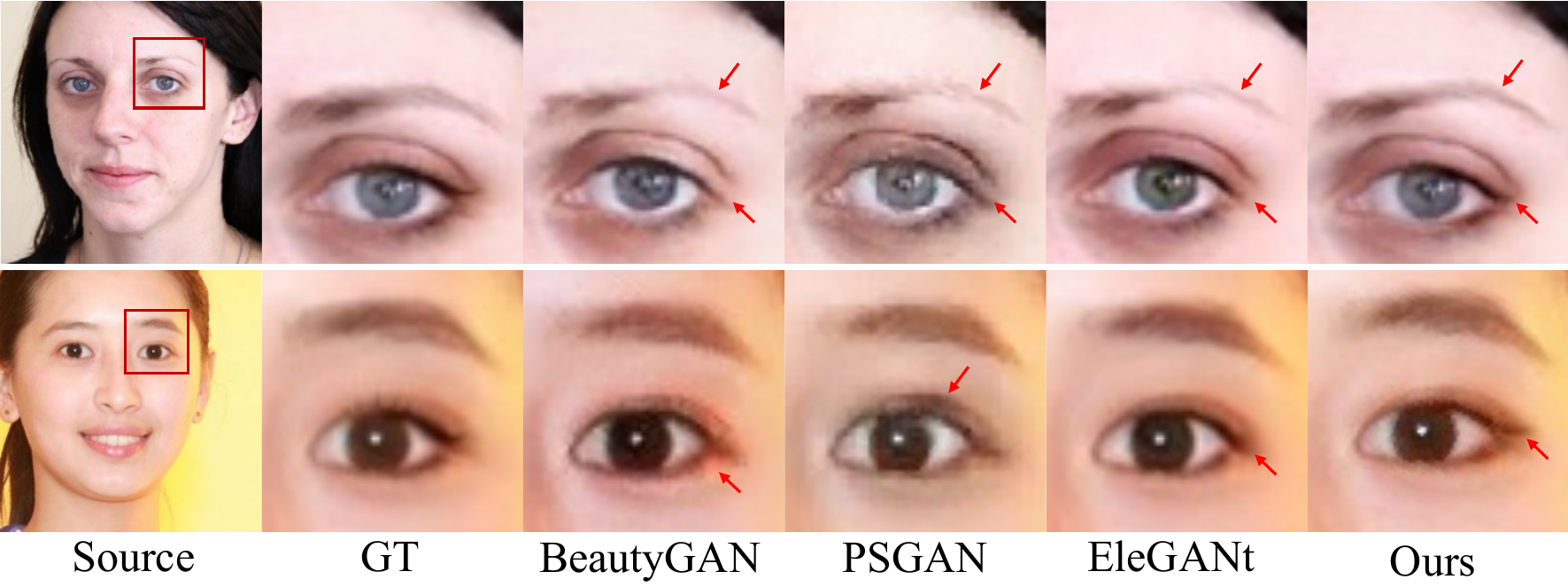}
        \caption{\textbf{Comparison of makeup details.} TinyBeauty generates finer details on eyeliner and complete eyebrows compared to previous methods. (Zoom in for more details.)}
        \label{fig:detail}
    \end{minipage}
    \hfill
    \begin{minipage}[t]{0.45\linewidth}
        \centering
        \includegraphics[width=\linewidth]{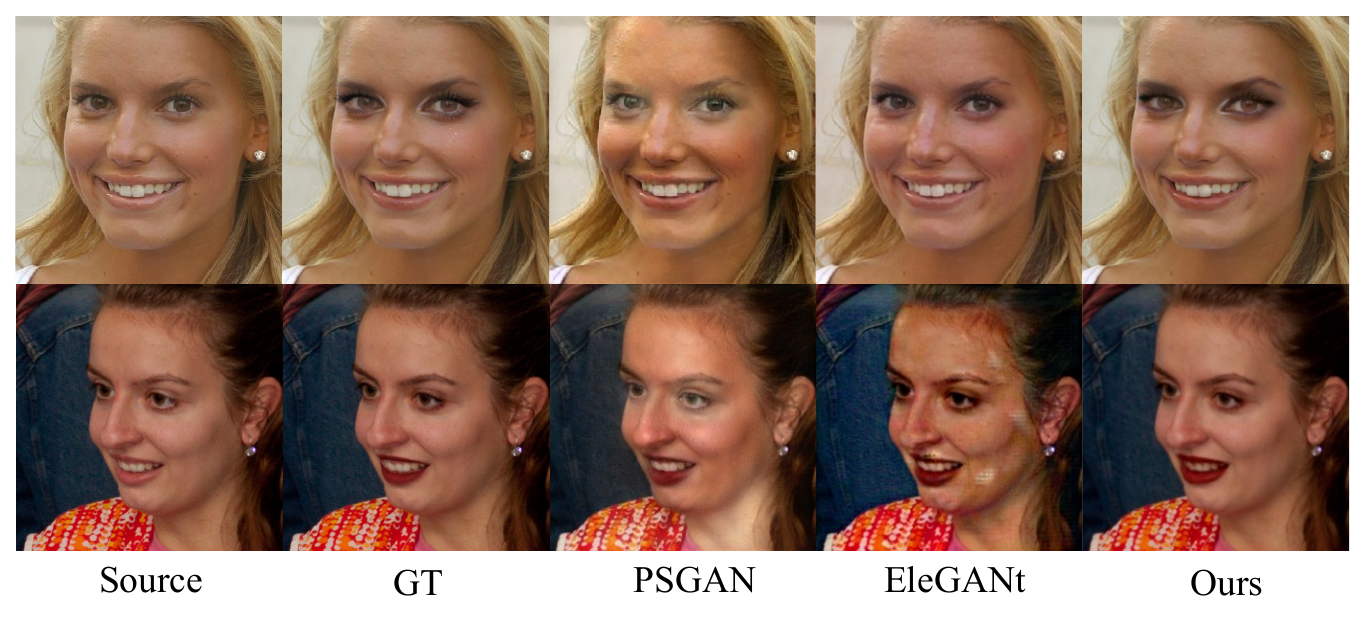}
        \caption{Visual comparison of the in-the-wild images with large poses and expressions.}
        \label{fig:inthewild}
    \end{minipage}
\end{figure}

To evaluate the effectiveness of our TinyBeauty on DDA-generated data, we conduct a comparison with several representative pre-trained facial makeup techniques, including BeautyGAN~\cite{BeautyGAN}, PSGAN~\cite{PSGAN}, SCGAN~\cite{SCGAN}, EleGANt~\cite{ELEGANT}, and BeautyREC~\cite{BeautyREC}. The experimental results, as depicted in~\cref{fig:res}, demonstrate the makeup data generated by our DDA and TinyBeauty are both visually pleasant and consistent with the ground-truth data. Beyond overall visual enhancement, TinyBeauty also provides auxiliary makeup functions such as automatic eyebrow completion and eyeliner application, as demonstrated in~\cref{fig:detail}. The precision of the makeup results stems from the combination of high-quality DDA-generated data and our targeted constraints on eyeliner design.

\begin{table}[tb]
    \centering
    \begin{minipage}{0.48\textwidth}
        \centering
        \caption{Parameters of TinyBeauty and competing methods. (+) means the method uses facial pre-processing including face parsing$^*$, and (-) means the method only has a single model.}
        \resizebox{\textwidth}{!}{
            \begin{tabular}{cccc}
                \toprule
                Method & Param.(M)$\downarrow$ & FLOPs(G)$\downarrow$ & RunTime(ms)$\downarrow$ \cr
                \cmidrule(lr){1-4}
                Face Parsing$^*$~\cite{faceparsing} &12.68 & 1.61& 13.29 \cr
                \cmidrule(lr){1-4}
                BeautyGAN~\cite{BeautyGAN} & 8.04 &24.70 & \underline{27.89} (-)\cr
                PSGAN~\cite{PSGAN} & 8.41 & 91.28 &  N/A (+)\cr
                SCGAN~\cite{SCGAN} & 35.33 & 288.51& 195.61 (+)\cr
                EleGANt~\cite{ELEGANT} & 10.27 & 127.94 & N/A (+)\cr
                BeautyREC~\cite{BeautyREC} & \underline{0.99} & \underline{12.58} & 206.46 (+)\cr
                \cmidrule(lr){1-4}
                \textbf{TinyBeauty} & \textbf{0.08} & \textbf{0.69} &  \textbf{2.18}  (-)\cr
                \bottomrule
            \end{tabular}
        }
        \label{tab:FLOP}
    \end{minipage}
    \hfill
    \begin{minipage}{0.48\textwidth}
        \centering
        \caption{The voting result ($\%$) of user study on test data. Participants are asked to rank the makeup quality based on accurate makeup color, clear details like eyebrows or eyeliner, pleasing effects, and minimal distortion. }
        \resizebox{\textwidth}{!}{
            \begin{tabular}{ccccccc}
                \toprule
                Method & Rank-1 & Rank-2 & Rank-3 & Rank-4 & Rank-5 & Rank-6 \cr
                \cmidrule(lr){1-7}
                BeautyGAN~\cite{BeautyGAN}          & 0.18    & 0.55    & 0.98    & 4.59    & 14.14   & \colorbox{lightgray}{\textbf{79.56}} \cr
                PSGAN~\cite{PSGAN}              & 1.56    & 0.86    & 11.43   & 6.80    & \colorbox{lightgray}{\textbf{64.75}}   & 13.98 \cr
                SCGAN~\cite{SCGAN}  & 1.00    & 1.76    & 27.71   & \colorbox{lightgray}{\textbf{54.27}}   & 13.76   & 1.46 \cr
                EleGANt~\cite{ELEGANT}            & 10.46   & \colorbox{lightgray}{\textbf{84.07}}   & 2.27    & 0.23    & 1.17    & 2.57 \cr
                BeautyREC~\cite{BeautyREC}& 0.28    & 1.95    & \colorbox{lightgray}{\textbf{55.06}}   & 34.13   & 6.15    & 2.43 \cr
                \cmidrule(lr){1-7}
                \textbf{TinyBeauty}                 & \colorbox{lightgray}{\textbf{86.56}}   & 10.81   & 2.55    & 0.05    & 0.03    & 0.00 \cr
                \bottomrule
            \end{tabular}
        }
        \label{tab:userstudy}
    \end{minipage}
\end{table}

\textbf{Quantitative Comparison.} We conduct a comprehensive evaluation by comparing TinyBeauty not only with the pre-trained models provided by earlier studies~\cite{BeautyGAN,BeautyREC,PSGAN,SCGAN,ELEGANT}, but also by retraining the two most advanced models~\cite{ELEGANT, BeautyREC} from those studies using our DAL. The comparative analysis, detailed in \cref{tab:PSNR}, reveals TinyBeauty's performance over the pre-established and newly trained models in terms of PSNR, FID, and LPIPS across both the FFHQ and MT datasets. Remarkably, despite its significantly smaller network size, TinyBeauty achieves comparable results to EleGANt in the same training set. The success lies in the high-quality paired data we generate, which enables even small networks to learn and apply stable makeup styles effectively. 
Moreover, our specialized design incorporating residual learning not only diminishes noise interference but also enhances TinyBeauty's performance, even surpassing larger predecessors. 
Notably, the MT dataset is not utilized during training, yet TinyBeauty yields a PSNR improvement of 1.49dB over prior methods on MT. 

\textbf{Model Efficiency Evaluation.} As shown in~\cref{tab:FLOP}, we measure FLOPs, parameter size, and inference time of each model. The models are packaged with CoreML~\cite{CoreML} and tested on an iPhone13 without including face pre-processing time. We also test a face parsing model~\cite{faceparsing} on iPhone13 for reference, commonly utilized by other approaches but not integrated within our TinyBeauty method.
Our model achieves an inference time of only 2.18ms for a 256x256 image, which is $13\times$ faster than the fastest competing method on iPhone13, i.e., BeautyGAN~\cite{BeautyGAN}. Specifically, TinyBeauty is approximately 6 times faster than face parsing on iPhone hardware, which is used as a pre-processing step in previous makeup transfer pipelines~\cite{BeautyREC, SCGAN, ELEGANT, PSGAN}. In contrast, our TinyBeauty only utilizes a single model by simplifying the optimization problem facilitated by amplified paired data. These results highlight TinyBeauty's efficiency and mobile deployment readiness, achieved by implementing our DAL scheme.

\begin{figure}[t]
  \centering
   \includegraphics[width=0.9\linewidth]{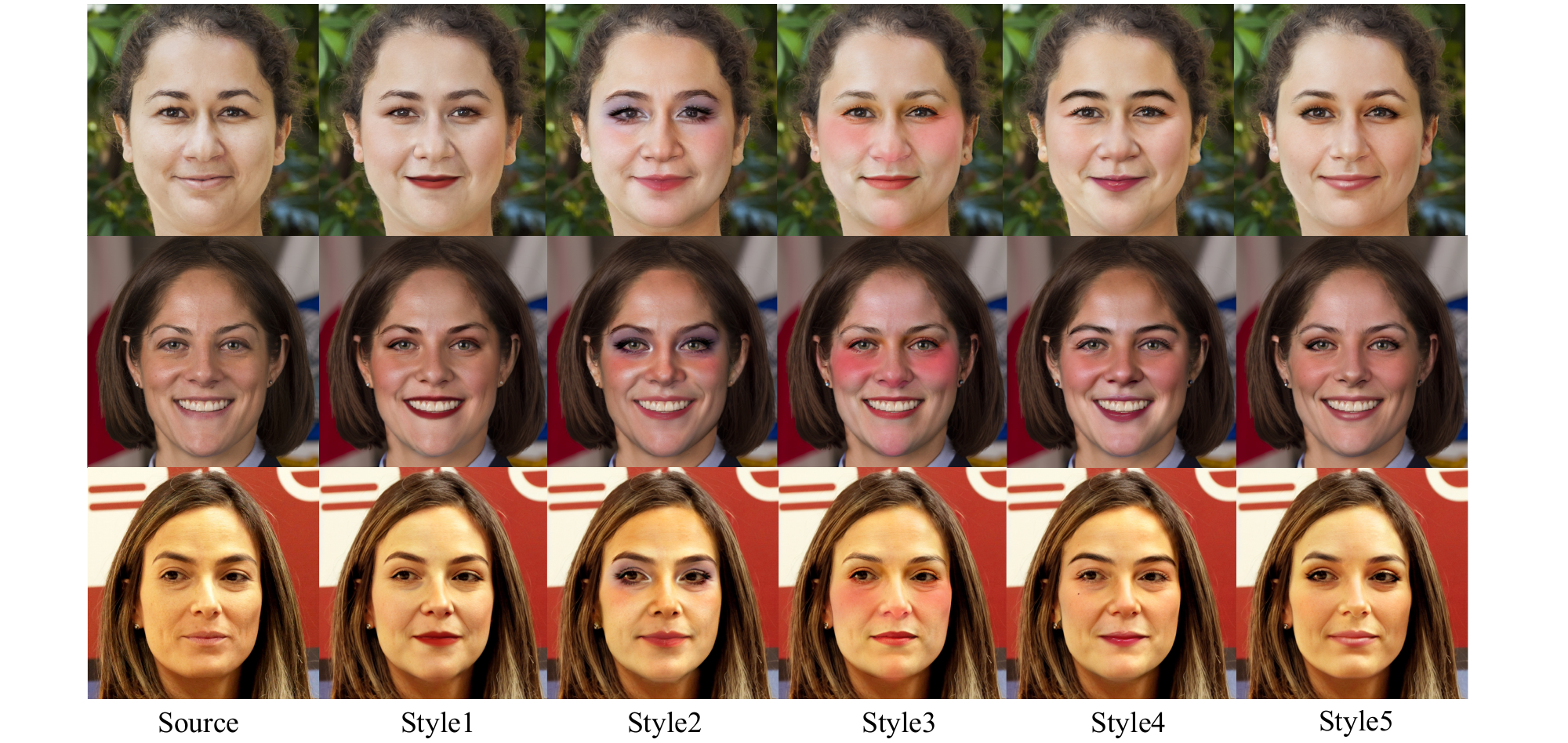}
   \caption{\textbf{Results of the diffusion-based data amplifier.} Consistent makeup styles are generated while retaining the facial content and identity of the original image.}
   \label{fig:diffusion_result}
\end{figure}

\begin{figure*}[t]
  \centering
   \includegraphics[width=0.95\linewidth]{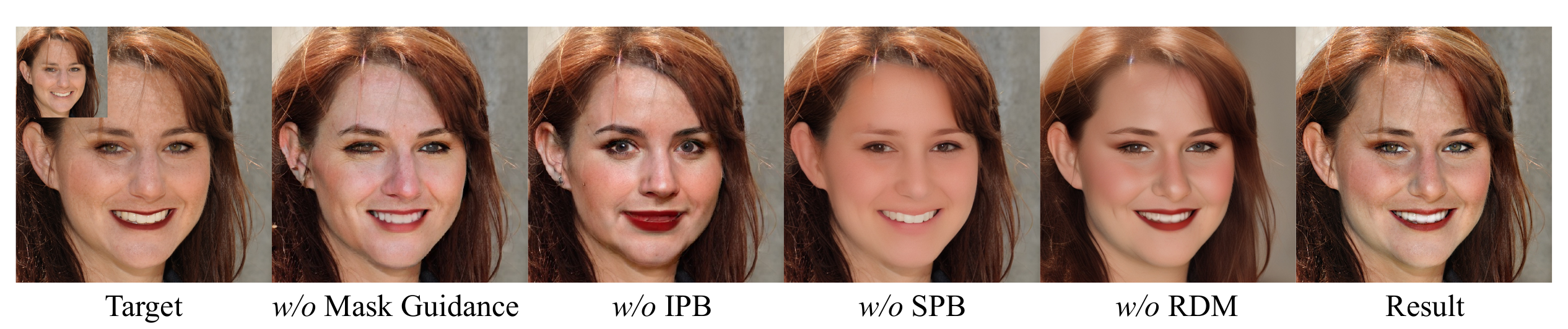}
   \caption{\textbf{Ablation study of RDM and modules in FMM of DDA.}}
   \label{fig:diffusion-abla}
\end{figure*}

\textbf{User Study.} 
We conduct a subjective evaluation on the FFHQ and MT test sets. Workers are shown makeup results from TinyBeauty and competing methods for raw images in random order. A total of 100 validated workers aged 19-48 from different cultural backgrounds participate in the fair and randomized evaluation. The average rankings are summarized in~\cref{tab:userstudy}.

\subsection{Ablation Study}
\textbf{Ablation Study of Diffusion-based Data Amplifier.} 
DDA is the core component of DAL. Its performance directly affects the ability of DAL model training.
The effectiveness of each module in the data amplifier is verified through combination experiments as shown in~\cref{fig:diffusion-abla}.
The experimental findings indicate that the integration of mask guidance and the IPB enables the DDA to more precisely retain the subject's identity. Simultaneously, the SPB delivers more accurate guidance on the makeup style, while the RDM maintains the facial texture, thereby enhancing the overall realism of the image. The combination of these modules enables DDA to produce high-quality makeup images.

\begin{figure}[t]
  \centering
   \includegraphics[width=0.85\linewidth]{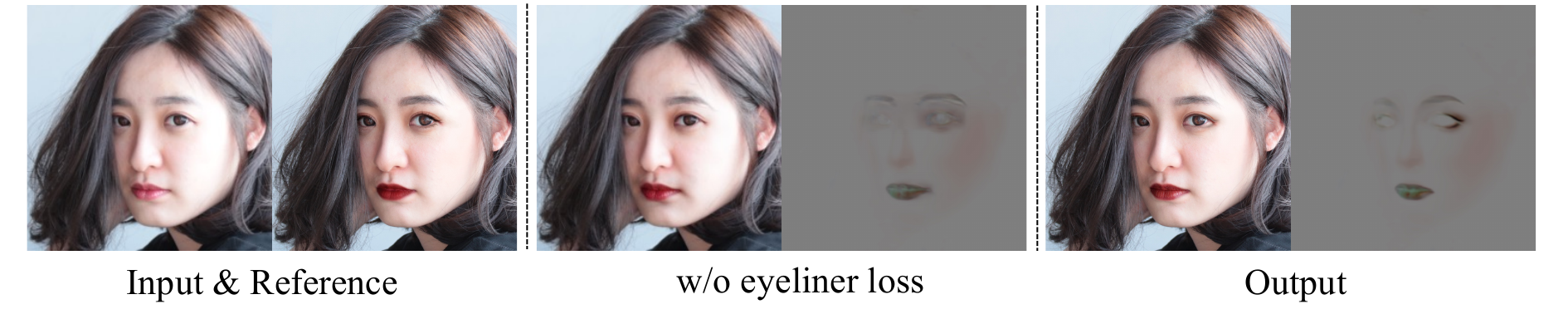}
   \caption{\textbf{Ablation study of eyeliner loss in TinyBeauty.} }
   \label{fig:tinym}
\end{figure}

\textbf{Component Analysis of TinyBeauty.} 
To investigate the effectiveness of eyeliner loss, we conduct ablation studies. As shown in~\cref{fig:tinym}, we find that removing the eyeliner loss on the region around the eyes indicates that it indeed greatly promotes learning of clear eyeliner.
This improvement is attributed to the fact that the network struggles to capture high-frequency signals such as eyeliner without specific constraints. By employing edge constraints to direct the network's learning process, our method enables it to discern and reproduce eyeliner details, which is not achievable by previous techniques.

\subsection{Results on the in-the-wild images and videos}
To further test the robustness of TinyBeauty, supplementary tests are conducted on both in-the-wild large-pose images and in-the-wild videos. The final makeup results are depicted in~\cref{fig:inthewild}.
Thanks to the large amount of paired data introduced by DAL, TinyBeauty has achieved significantly better performance than previous methods such as PSGAN~\cite{PSGAN} and EleGANt~\cite{ELEGANT}, which specializes in addressing facial makeup with large poses and expressions. Additionally, TinyBeauty demonstrates stronger makeup application abilities in videos, as shown in Supplementary Material.

\section{Conclusion}
In this paper, we introduce a revolutionary approach to facial makeup application with the development of Data Amplify Learning, and the implementation of a compact makeup model, TinyBeauty. By leveraging the innovative Residual Diffusion Model and Fine-Grained Makeup Module within our Data Amplifier, we effectively amplify limited paired data as the training data of TinyBeauty, a tiny 14-layer convolutional model replacing previous cumbersome pipelines. This results in a swift 2.18ms mobile makeup application and a 17.3\% PSNR quality boost, establishing a new milestone for real-time, low-resource mobile makeup.

\section*{Acknowledgements}
 This work was supported by National Science Foundation of China (U20B2072, 61976137). This work was also partly supported by SJTU Medical Engineering Cross Research Grant YG2021ZD18.
%
%
\bibliographystyle{splncs04}
\bibliography{main}
\end{document}


\title{Toward Tiny and High-quality Facial Makeup with Data Amplify Learning} 

\titlerunning{Abbreviated paper title}

\author{First Author\inst{1}\orcidlink{0000-1111-2222-3333} \and
Second Author\inst{2,3}\orcidlink{1111-2222-3333-4444} \and
Third Author\inst{3}\orcidlink{2222--3333-4444-5555}}

\authorrunning{F.~Author et al.}

\institute{Princeton University, Princeton NJ 08544, USA \and
Springer Heidelberg, Tiergartenstr.~17, 69121 Heidelberg, Germany
\email{lncs@springer.com}\\
\url{http://www.springer.com/gp/computer-science/lncs} \and
ABC Institute, Rupert-Karls-University Heidelberg, Heidelberg, Germany\\
\email{\{abc,lncs\}@uni-heidelberg.de}}

\makesupptitle
\section{Supplementary Results}
The anonymous homepage for our project can be found  \href{https://anonymous.4open.science/w/TinyBeauty-7A4F/}{here}.

\subsection{Supplementary results of diffusion-based data amplifier}
\begin{figure*}[t]
  \centering
   \includegraphics[width=0.9\linewidth]{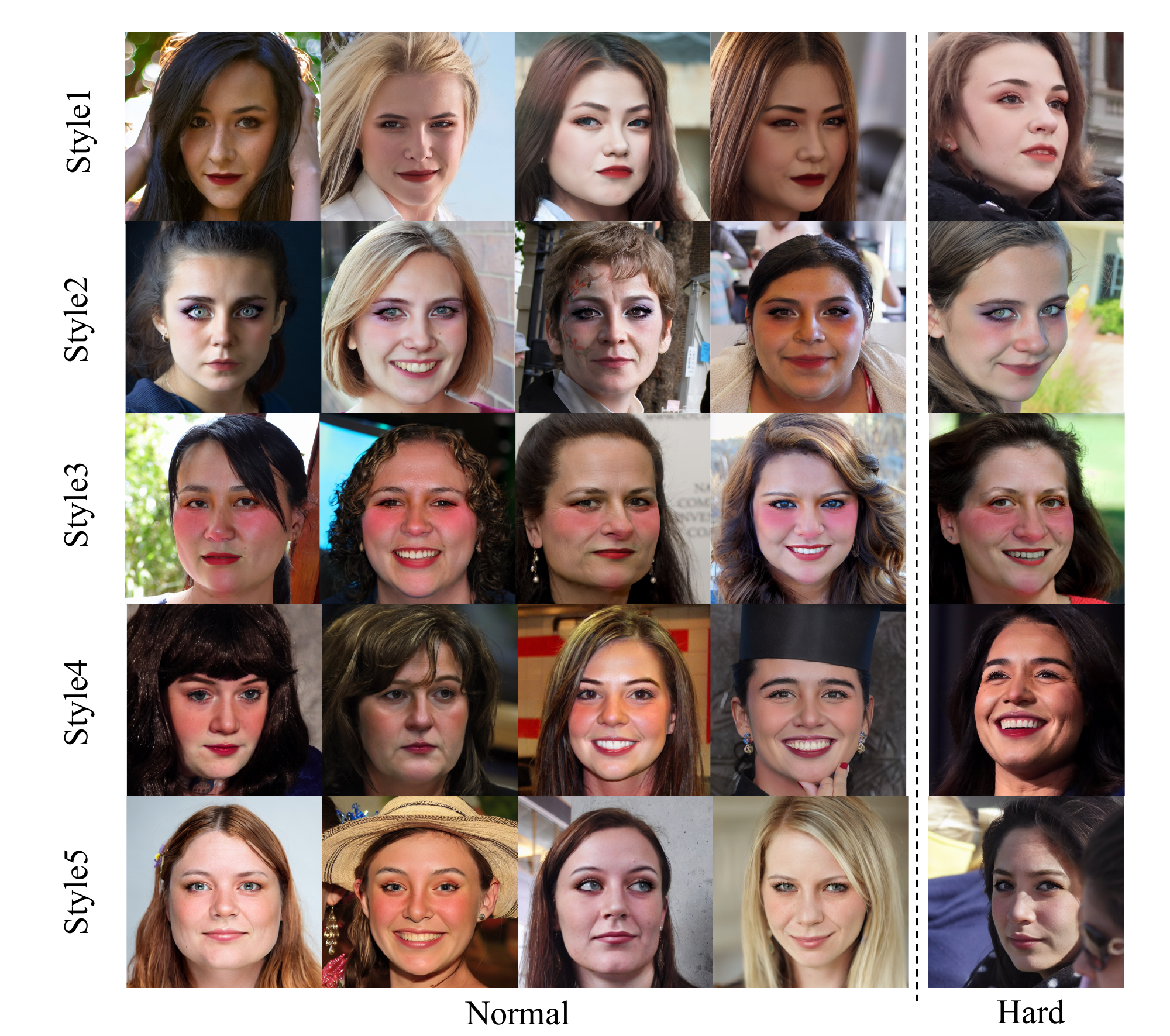}
    \caption{\textbf{Additional results generated by diffusion-based data amplifier.}}
   \label{fig:diffusion}
\end{figure*}

As illustrated in~\cref{fig:diffusion}, the diffusion-based data amplifier can learn five distinct makeup styles, each with its characteristic LoRA. It can apply makeup styles with stable and high quality to non-makeup images while retaining the facial features of the original portrait.

 \begin{figure*}[hb]
  \centering
   \includegraphics[width=1.0\linewidth]{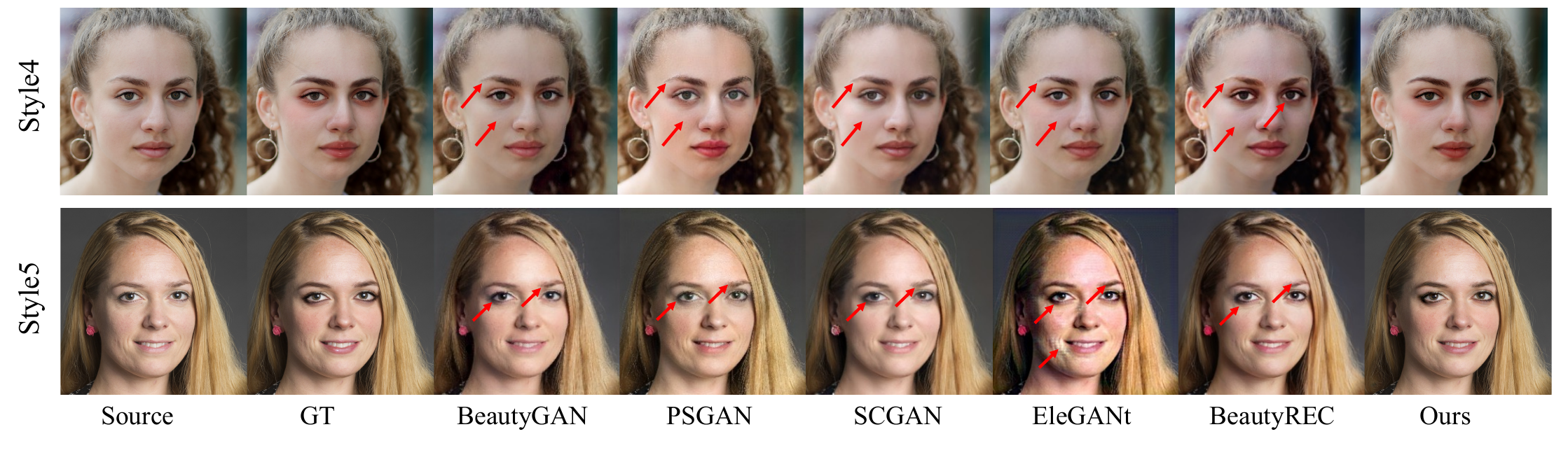}
    \caption{\textbf{Visual comparison of TineBeauty and competing methods on the FFHQ~\cite{FFHQ} image, in Style4 and Style5.}}
   \label{fig:FFHQ_res}
\end{figure*}

\subsection{Supplementary facial makeup results on the FFHQ dataset}
 Results from our lightweight makeup model and competing methods for Style4 and Style5 on the FFHQ~\cite{FFHQ} dataset are displayed in~\cref{fig:FFHQ_res}. Our method exhibits notable advantages over prior approaches along three primary dimensions. Firstly, makeup around the eye region is applied with heightened precision by our TinyBeauty. Secondly, TinyBeauty possesses the capability to color and define eyebrow regions realistically. Thirdly, delicate cosmetic effects such as blush and contouring are delineated with accurate localization.

\begin{figure*}[h]
  \centering
   \includegraphics[width=1.0\linewidth]{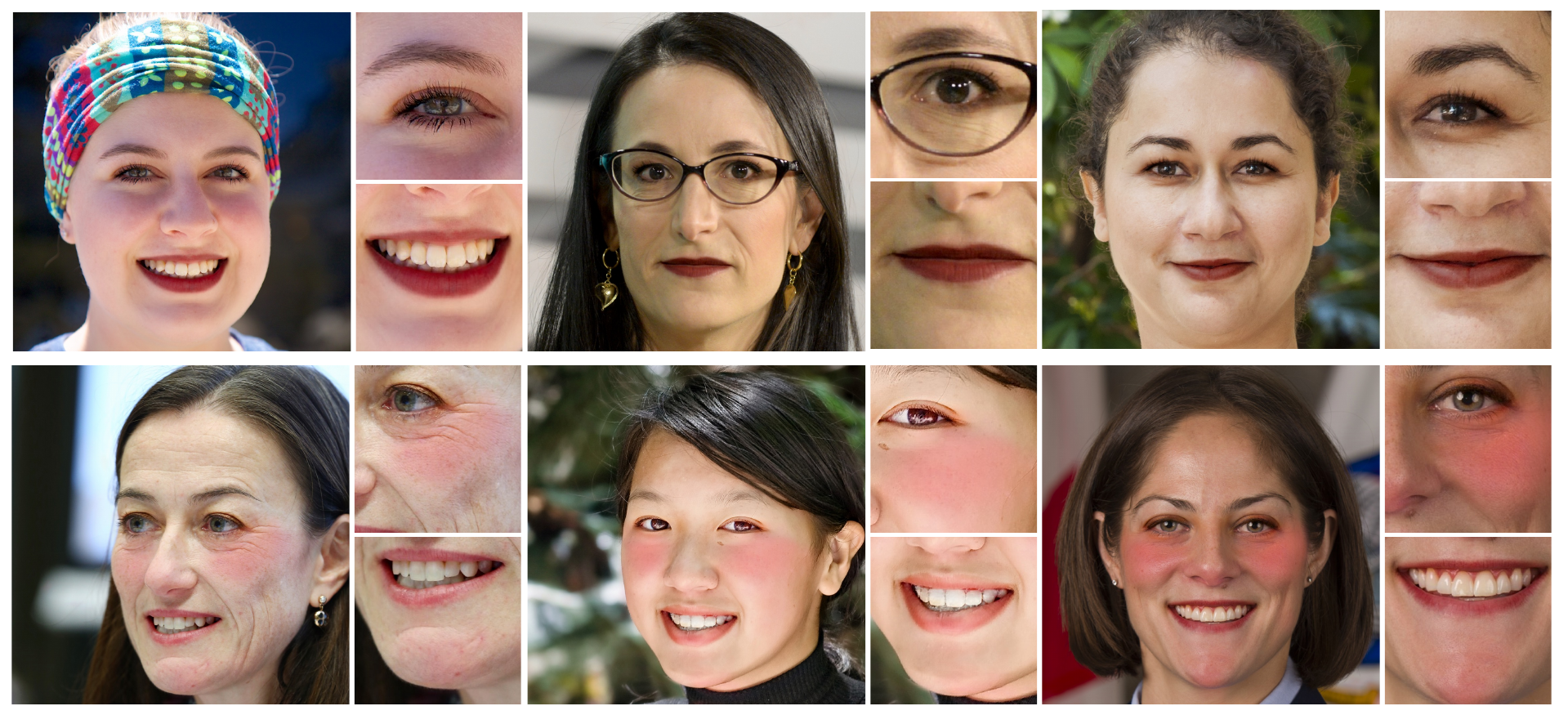}
    \caption{\textbf{Facial makeup results on high-resolution (1024$\times$1024) images.}}
   \label{fig:HDR}
\end{figure*}

\subsection{Supplementary facial makeup results on the MT dataset}
\begin{figure*}[htb]
  \centering
   \includegraphics[width=1.0\linewidth]{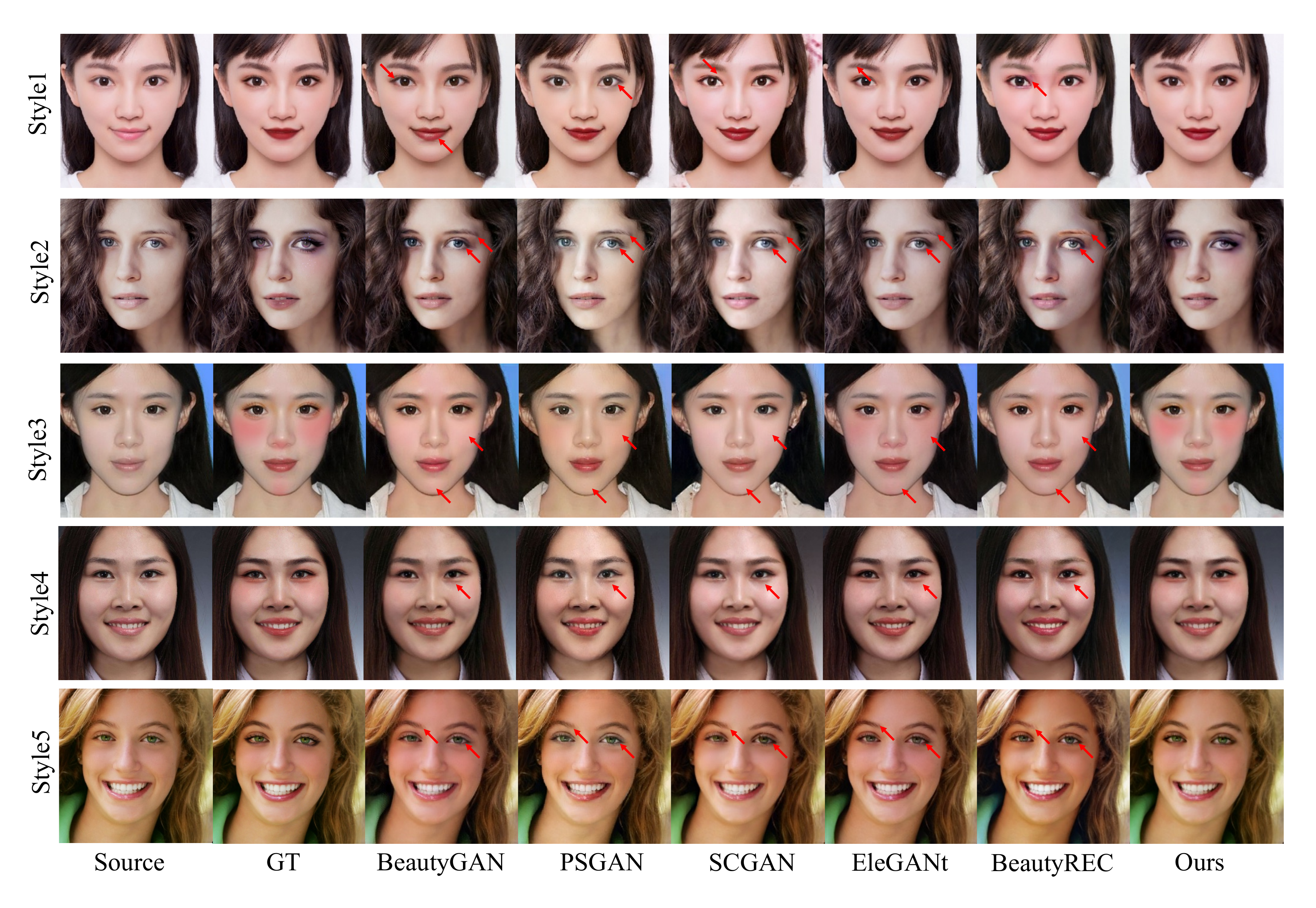}
    \caption{\textbf{Visual comparison of TineBeauty and competing methods on the MT~\cite{BeautyGAN} images.}}
   \label{fig:MT_res}
\end{figure*}

While our model training utilized solely the FFHQ dataset due to the constrained size of the MT dataset, we performed supplementary testing of our FFHQ-trained model on the MT dataset, as depicted in~\cref{fig:MT_res}, to conduct a fairer comparative assessment of TinyBeauty against preceding techniques. Observation of the results indicates that regardless of the exclusion of the MT dataset from training, the generated makeup outputs produced by our approach still surpass those of previous methods in visual quality. This suggests that the representation of facial attributes learned from FFHQ enabled our solution to more precisely depict makeup allocation, demonstrating its ability to generalize to other face image domains beyond the dataset it was exclusively trained on.

\subsection{Facial makeup results on high-resolution images.}
Prior generative networks adopted an approach with direct mapping from input non-makeup portraits to output makeup depictions. This paradigm necessitates substantial computational costs for high-resolution cosmetic transfer due to processing the entire face image. Additionally, such transformations may induce unintended alterations to facial content properties during image synthesis. To circumvent these limitations, our lightweight model is formulated to yield residual representations specific to sole makeup attributes rather than fully reconstructed portraits. 

\begin{figure*}[h]
  \centering
   \includegraphics[width=1.0\linewidth]{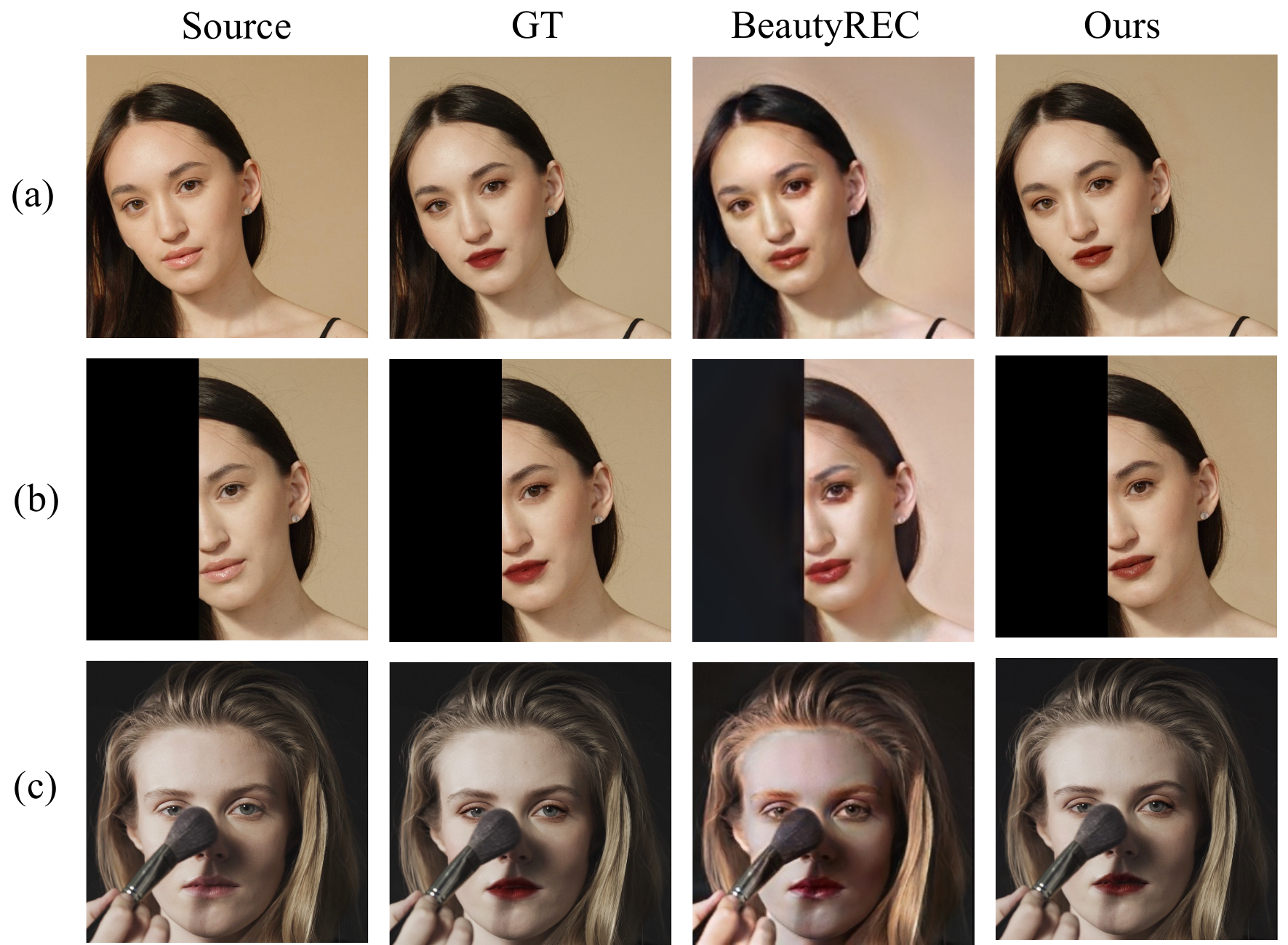}
    \caption{\textbf{Visual comparison of TineBeauty and BeautyREC~\cite{BeautyREC} on challenging out-of-distribution examples: (a) a non-ideal centered face, (b) a face partially captured outside the bounds of the image, and (c) an occluded face.}}
   \label{fig:hard}
\end{figure*}

\begin{table}[b]
        \centering
        \setlength\tabcolsep{3pt}
        \fontsize{8}{12}\selectfont    
        \caption{Results of TinyBauty and competing methods on FFHQ and MT datasets, in Style2.}
        \begin{tabular}{ccccccc}
                \toprule
                \toprule
                \multirow{Method}&
                \multicolumn{3}{c}{FFHQ} &
            \multicolumn{3}{c}{MT}\cr
                \cmidrule(lr){2-4}  
                \cmidrule(lr){5-7}  
                & PSNR$\uparrow$ & FID$\downarrow$ & LPIPS$\downarrow$ & PSNR$\uparrow$ & FID$\downarrow$ & LPIPS$\downarrow$ \cr
            \cmidrule(lr){1-7}
                BeautyGAN~\cite{BeautyGAN} & 25.95 & 45.38 & 0.0595 & 25.84 & 31.44 & 0.0537 \cr
                PSGAN~\cite{PSGAN}     & 25.31 & 35.88 & 0.0624 & 26.99 & 18.31 & 0.0419 \cr
                SCGAN~\cite{SCGAN}     & 26.84 & 38.24 & 0.0527 & 25.61 & 35.96 & 0.0572 \cr
                EleGANt~\cite{ELEGANT}   & 29.34 & 26.16 & 0.0441 & 30.47 & 13.83 & 0.0296 \cr
                BeautyREC~\cite{BeautyREC} & 24.09 & 28.32 & 0.0591 & 25.46 & 23.24 & 0.0508 \cr

            \cmidrule(lr){1-7}
            \textbf{TinyBeauty} & \textbf{33.331} & \textbf{10.327} & \textbf{0.0258} & \textbf{32.445} & \textbf{11.313} & \textbf{0.0335} \cr
                \bottomrule
                \bottomrule
        \end{tabular}\vspace{0cm}
        \label{tab:1}
\end{table}

\begin{table}
        \centering
        \setlength\tabcolsep{3pt}
        \fontsize{8}{12}\selectfont    
        \caption{Results of TinyBauty and competing methods on FFHQ dataset and MT dataset, in Style3.}
        \begin{tabular}{ccccccc}
                \toprule
                \toprule
                \multirow{2}{*}{Method}&
                \multicolumn{3}{c}{FFHQ} &
            \multicolumn{3}{c}{MT}\cr
                \cmidrule(lr){2-4}  
                \cmidrule(lr){5-7}  
                & PSNR$\uparrow$ & FID$\downarrow$ & LPIPS$\downarrow$ & PSNR$\uparrow$ & FID$\downarrow$ & LPIPS$\downarrow$ \cr
                \cmidrule(lr){1-7}

                BeautyGAN~\cite{BeautyGAN} & 26.41 & 48.67 & 0.0645 & 27.51 & 47.55 & 0.0628 \cr
                PSGAN~\cite{PSGAN}     & 25.67 & 38.97 & 0.0677 & 27.71 & 38.10 & 0.0518 \cr
                SCGAN~\cite{SCGAN}     & 27.65 & 44.52 & 0.0541 & 27.17 & 53.16 & 0.0598 \cr
                EleGANt~\cite{ELEGANT}   & 29.91 & 30.55 & 0.0483 & 32.06 & 37.20 & 0.0368 \cr
                BeautyREC~\cite{BeautyREC} & 24.55 & 33.12 & 0.0626 & 26.62 & 42.22 & 0.0601 \cr

            \cmidrule(lr){1-7}
            \textbf{TinyBeauty} & \textbf{34.72} & \textbf{11.14} & \textbf{0.0288} & \textbf{33.788} & \textbf{12.029} & \textbf{0.0233} \cr
                \bottomrule
                \bottomrule
        \end{tabular}\vspace{0cm}
        \label{tab:2}
\end{table}

\begin{table}[ht]
        \centering
        \setlength\tabcolsep{3pt}
        \fontsize{8}{12}\selectfont    
        \caption{Results of TinyBauty and competing methods on FFHQ dataset and MT dataset, in Style4.}
        \begin{tabular}{ccccccc}
                \toprule
                \toprule
                \multirow{Method}&
                \multicolumn{3}{c}{FFHQ} &
            \multicolumn{3}{c}{MT}\cr
                \cmidrule(lr){2-4}  
                \cmidrule(lr){5-7}  
                & PSNR$\uparrow$ & FID$\downarrow$ & LPIPS$\downarrow$ & PSNR$\uparrow$ & FID$\downarrow$ & LPIPS$\downarrow$ \cr
                \cmidrule(lr){1-7}
            
                BeautyGAN~\cite{BeautyGAN} & 26.75 & 44.73 & 0.0568 & 27.65 & 28.96 & 0.0467 \cr
                PSGAN~\cite{PSGAN}     & 25.68 & 35.69 & 0.0607 & 27.94 & 14.49 & 0.0345 \cr
                SCGAN~\cite{SCGAN}     & 27.97 & 35.35 & 0.0453 & 27.29 & 30.68 & 0.0615 \cr
                EleGANt~\cite{ELEGANT}   & 30.05 & 24.18 & 0.0404 & 32.22 & 10.86 & 0.0200 \cr
                BeautyREC~\cite{BeautyREC} & 24.83 & 27.02 & 0.0546 & 27.00 & 19.78 & 0.0424 \cr

            \cmidrule(lr){1-7}
            \textbf{TinyBeauty} & \textbf{34.737} & \textbf{9.29} & \textbf{0.0225} & \textbf{33.658} & \textbf{10.756} & \textbf{0.0378} \cr
                \bottomrule
                \bottomrule
        \end{tabular}\vspace{0cm}
        \label{tab:3}
\end{table}

\begin{table}
        \centering
         \setlength\tabcolsep{3pt}
        \fontsize{8}{12}\selectfont    
        \caption{Results of TinyBauty and competing methods on FFHQ dataset and MT dataset, in Style5.}
        \begin{tabular}{ccccccc}
                \toprule
                \toprule
                \multirow{2}{*}{Method}&
                \multicolumn{3}{c}{FFHQ} &
            \multicolumn{3}{c}{MT}\cr
                \cmidrule(lr){2-4}  
                \cmidrule(lr){5-7}  
                & PSNR$\uparrow$ & FID$\downarrow$ & LPIPS$\downarrow$ & PSNR$\uparrow$ & FID$\downarrow$ & LPIPS$\downarrow$ \cr
                \cmidrule(lr){1-7}
            
                BeautyGAN~\cite{BeautyGAN} & 26.47 & 46.20 & 0.0617 & 27.21 & 28.87 & 0.0519 \cr
                PSGAN~\cite{PSGAN}     & 25.57 & 36.31 & 0.0646 & 27.70 & 18.47 & 0.0388 \cr
                SCGAN~\cite{SCGAN}     & 26.97 & 42.62 & 0.0572 & 26.60 & 31.32 & 0.0515 \cr
                EleGANt~\cite{ELEGANT}   & 29.88 & 27.27 & 0.0442 & 32.50 & 14.28 & 0.0266 \cr
                BeautyREC~\cite{BeautyREC} & 24.46 & 29.72 & 0.0590 & 26.38 & 22.45 & 0.0503 \cr

            \cmidrule(lr){1-7}
            TinyBeauty & \textbf{35.652} & \textbf{9.18} & \textbf{0.0143} & \textbf{34.684} & \textbf{9.28} & \textbf{0.0216} \cr
                \bottomrule
                \bottomrule
        \end{tabular}\vspace{0cm}
        \label{tab:4}
\end{table}

\begin{figure}[h]
  \centering
   \includegraphics[width=0.7\linewidth]{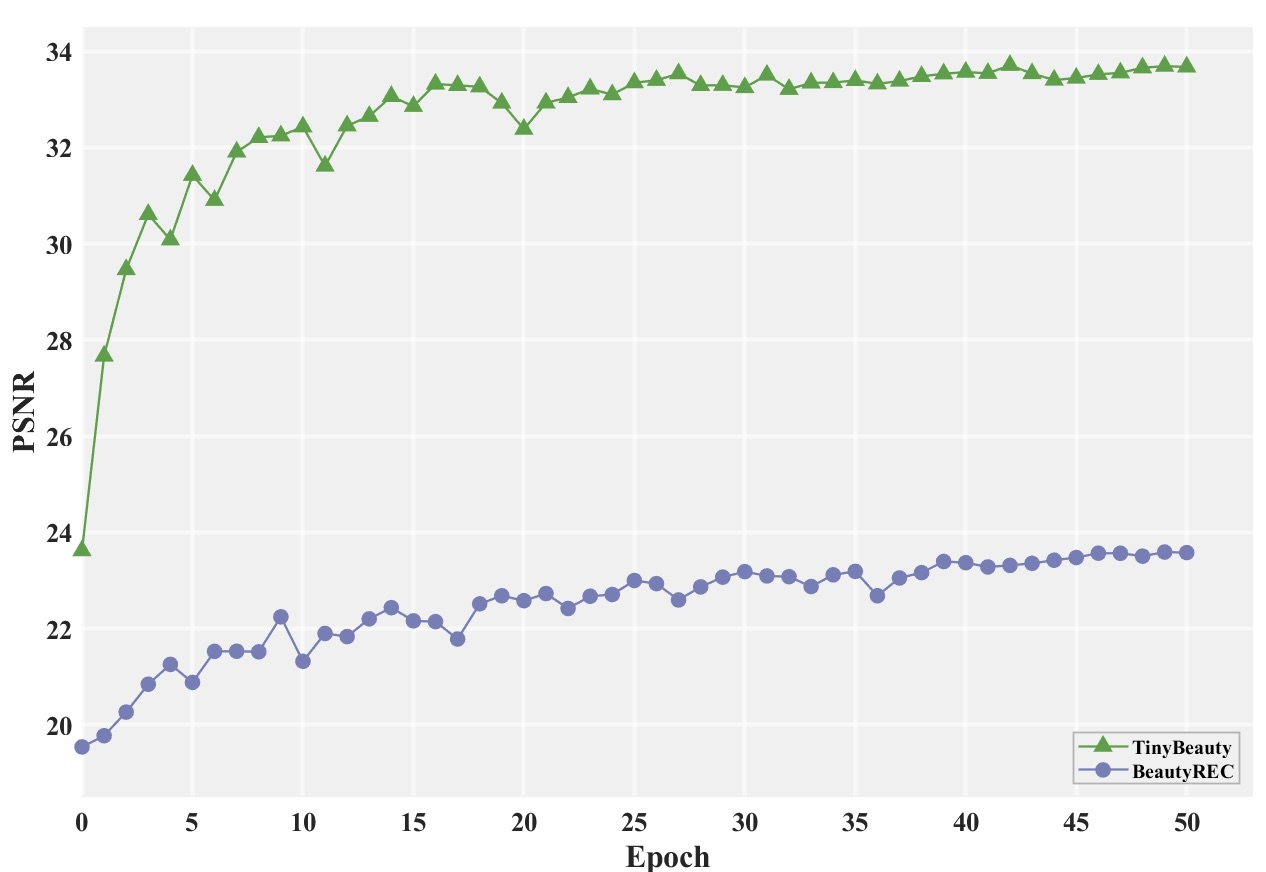}
    \caption{\textbf{Comparison of PSNR convergence curves between TinyBeauty and BeautyREC during training.}}
   \label{fig:opti}
\end{figure}

\begin{figure}[h]
  \centering
   \includegraphics[width=0.7\linewidth]{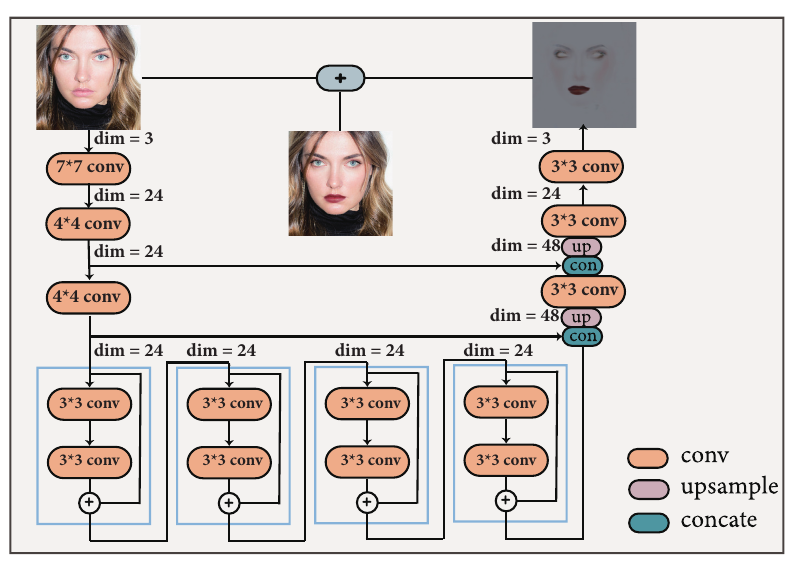}
    \caption{\textbf{Architecture of TinyBeauty Model.}}
   \label{fig:network}
   \vspace{-0.4cm}
\end{figure}

Specifically, prior techniques exclusively operated on facial portraits resized to 256$\times$256 pixels. To conduct makeup transfer at larger 1024$\times$1024 resolutions, their networks would either necessitate retraining with enlarged 1024$\times$1024 inputs, drastically increasing computational overhead, or rely on upsampling 256$\times$256 outputs to 1024x1024 pixels, incurring information loss. In contrast, our TinyBeauty model retains a 256$\times$256 input dimension, requiring only resizing its makeup residual layers to 1024$\times$1024 before reconstructing the final output by summation with the original 1024$\times$1024 non-makeup portrait. This avoids retraining demands and precludes the degradation in granularity associated with resizing low-resolution generation results. Critically, it maintains computational efficiency equal to 256$\times$256 processing since the primary network architecture undergoes no modification to accommodate higher resolutions.

\subsection{Facial makeup results on hard cases.}
To further evaluate TinyBeauty's tolerance to challenging input deviations, supplementary comparisons against BeautyREC~\cite{BeautyREC} were conducted on irregular example portraits. Situations explored included: (a) a non-ideal centered face, (b) a face partially captured outside the bounds of the image, and (c) an occluded face. These archetypes simulate varied real-world acquisition scenarios undermining standard assumptions.

As shown in~\cref{fig:hard}, TinyBeauty can successfully apply makeup without prerequisite face detection or localization pre-processing, unlike BeautyREC which failed to properly segment features. This validates TinyBeauty's end-to-end generation ability from raw inputs alone and TinyBeauty's superior tolerance to input variations.

\subsection{Supplementary quantitative results.}
In addition to the quantitative analyses on Style 1 reported in the primary manuscript, equivalent objective assessments were performed for the remaining four makeup styles. As depicted in~\cref{tab:1},~\cref{tab:2},~\cref{tab:3},~\cref{tab:4}, our approach achieves favorable performance as evaluated by PSNR, FID~\cite{FID}, and LPIPS~\cite{LPIPS} on both the FFHQ and MT datasets, surpassing prior cosmetic transfer techniques by a substantial margin according to each numerical metric. Our lightweight model consistently outperforms predecessors across all styles assessed and datasets, as evidenced by higher PSNR and lower FID and LPIPS scores. These results substantiate TinyBeauty's demonstrated superior makeup simulation ability and position it as the current leading solution among facial makeup models.

\subsection{Comparison of convergence curves.}
To further validate that TinyBeauty's training procedure is simpler than previous methods, we compared the rising PSNR curves of TinyBeauty and BeautyREC~\cite{BeautyREC} during 50 epochs of training on 200 images. Specifically, as shown in~\cref{fig:opti}, TinyBeauty approached convergence within 15 training iterations, while BeautyREC did not approach convergence until 40 iterations. Additionally, TinyBeauty's final PSNR score far exceeded that of BeautyREC. These optimization results quantitatively demonstrate that TinyBeauty requires substantially fewer parameter updates than BeautyREC but reaches both fast convergence and superior performance, benefiting from direct L1 loss.

\section{Additional Losses}
\textbf{Perceptual Similarity Loss.} To maintain visual fidelity between the input face image and synthesized makeup result, we introduce a perceptual similarity loss $\mathcal{L}_{per}$ following previous works~\cite{BeautyGAN, BeautyREC, PSGAN}. This loss measures the Euclidean distance in feature space between the conv4 layer activations of the input and output images when fed through a pre-trained VGG-19 network~\cite{VGG19}:
\begin{equation}
\mathcal{L}_{per} = ||\phi_{vgg}(x) - \phi_{vgg}(y')||_2,
\end{equation}
where $\phi_{vgg}$ denotes the convolutional features extracted from the conv4 layer before activation. By minimizing this loss, our model generates makeup results that align with the input face image semantically. 

\textbf{Adversarial Loss.}
To help the makeup model generate realistic makeup outputs, we incorporate global and local adversarial losses to further constrain the makeup outputs. The ultimate formulation of the adversarial loss is:
\begin{equation}
    \mathcal{L}_{adv} = \mathcal{L}_{adv}^{global} + \mathcal{L}_{adv}^{eyes} +\mathcal{L}_{adv}^{eyebrows}+\mathcal{L}_{adv}^{skin}+\mathcal{L}_{adv}^{lips},
\end{equation}
where $\mathcal{L}{adv}^{global}$ penalizes deviations from the ground truth makeup style globally across the entire facial region. The localized losses $\mathcal{L}{adv}^{eyes}$, $\mathcal{L}{adv}^{eyebrows}$, $\mathcal{L}{adv}^{skin}$, and $\mathcal{L}_{adv}^{lips}$ further optimize consistency specifically within semantically meaningful regions. Through this formulation, we aim to refine the makeup transfer at both global and localized levels, enforcing fine-grained fidelity to the target style distribution over key facial components.

\textbf{Total Loss.} The total loss is a combination of the above-mentioned losses and losses in the main paper, which can be expressed as:
\begin{equation}
\mathcal{L}_{total} = \mathcal{L}_{rec} + \mathcal{L}_{s} + \lambda_{per} \mathcal{L}_{per} + \lambda_{adv} \mathcal{L}_{adv}, 
\end{equation}
where $\lambda_{per} = 0.005$ and $\lambda_{adv} = 0.5$ are the corresponding weights for balancing the magnitudes of losses.

\section{Structure of TinyBeauty Model}
To show the network structure of our TinyBeauty model more clearly, we drew the structural configurations of the convolutional layers comprising the network, as shown in~\cref{fig:network}. Particularly, we employ ReLU as an activation layer after each convolutional stage for nonlinear mapping, alongside Instance Normalization to regulate the distribution of input features. By leveraging such operation-efficient modules, our model achieves markedly expedited inference amenable to deployment on mobile devices with constrained computational budgets.

\bibliographystyle{splncs04}
\bibliography{main}